%% file: main.tex
\documentclass[twoside]{article}

\usepackage[accepted]{aistats2024}
\usepackage{amsmath,amssymb,amsfonts,amsthm}
\usepackage{algorithmic}
\usepackage{graphicx}
\usepackage{textcomp}
\usepackage[ruled,linesnumbered]{algorithm2e}
\usepackage{xcolor}
\usepackage{subcaption}
\usepackage[round]{natbib}
\usepackage{lipsum}
\usepackage{url}

\newcommand{\wes}[1]{{\leavevmode\color{black}#1}}
\newcommand{\siva}[1]{{\leavevmode\color{black}#1}}

\newcommand{\mc}[1]{\mathcal{#1}}

\theoremstyle{plain}

\newtheorem{definition}{Definition}

\newtheorem{assumption}{Assumption}
\newtheorem{proposition}{Proposition}
\newtheorem{theorem}{Theorem}
\newtheorem{corollary}{Corollary}
\newtheorem{lemma}{Lemma}
\theoremstyle{remark}
\newtheorem{remark}{Remark}

\newcommand{\bR} { {\mathbb R}}
\newcommand{\norm}[1]{ \left\Vert #1 \right\Vert}

\begin{document}

%
\runningtitle{Sampling-based Safe Reinforcement Learning for Nonlinear Dynamical Systems}

%
\runningauthor{W. A. Suttle, V. K. Sharma, K. C. Kosaraju, S. Sivaranjani, J. Liu, V. Gupta, B. M. Sadler}


\twocolumn[
\aistatstitle{Sampling-based Safe Reinforcement Learning for \\Nonlinear Dynamical Systems}

\aistatsauthor{ Wesley A. Suttle \And Vipul K. Sharma \And Krishna C. Kosaraju }

\aistatsaddress{ U.S. Army Research \\ Laboratory \And Purdue University \And Clemson University } 

\aistatsauthor{S. Sivaranjani \And Ji Liu \And Vijay Gupta \And Brian M. Sadler }

\aistatsaddress{ Purdue University \And Stony Brook University \And Purdue University \And U.S. Army Research \\ Laboratory }]

\begin{abstract}
    We develop provably safe and convergent reinforcement learning (RL) algorithms for control of nonlinear dynamical systems, bridging the gap between the hard safety guarantees of control theory and the convergence guarantees of RL theory. Recent advances at the intersection of control and RL follow a two-stage, safety filter approach to enforcing hard safety constraints: model-free RL is used to learn a potentially unsafe controller, whose actions are projected onto safe sets prescribed, for example, by a control barrier function. Though safe, such approaches lose any convergence guarantees enjoyed by the underlying RL methods. In this paper, we develop a single-stage, sampling-based approach to hard constraint satisfaction that learns RL controllers enjoying classical convergence guarantees while satisfying hard safety constraints throughout training and deployment. We validate the efficacy of our approach in simulation, including safe control of a quadcopter in a challenging obstacle avoidance problem, and demonstrate that it outperforms existing benchmarks.
\end{abstract}

\input{AISTATS2024/sections/Introduction}
\input{AISTATS2024/sections/Problem_Setting}
\input{AISTATS2024/sections/Theoretical_Results}
\input{AISTATS2024/sections/Experimental_Results}
\input{AISTATS2024/sections/Conclusion}
\input{AISTATS2024/sections/Acknowledgements}

\bibliographystyle{plainnat} 
\bibliography{main}


\newpage

\onecolumn
\appendix

\input{AISTATS2024/supplement}
\end{document}

%% file: AISTATS2024/sections/Introduction.tex
\section{INTRODUCTION}

Learning-based methods for safe control of physical systems have been gaining increasing attention \citep{brunke2022safe}.
%
%
RL is especially powerful for the control of systems where performance feedback in the form of a scalar reward is available, but the dynamics are unknown \citep{sutton2018reinforcement}. In such settings, RL methods can learn a controller maximizing reward through direct interaction with the environment.
However, due to physical realities such as the need to guarantee safety, practical application of RL to control of physical systems requires constraints on the control policies throughout training \citep{garcia2015comprehensive}. While directly constraining the action space to a static, narrowly defined set of ``safe'' actions is frequently employed in practice, this can lead to learning highly suboptimal policies and more nuanced methods are therefore required. Furthermore, in most physical systems it is non-trivial to directly translate complex safety constraints on the states into allowable actions.

A variety of RL approaches to the problem of safe learning for control have been proposed in the literature (see \cite{brunke2022safe} for a comprehensive survey), including RL methods for safety-focused problems formulated as constrained Markov decision processes (CMDPs) \citep{altman2021constrained}, methods for learning to achieve safety through stability \citep{berkenkamp2017safe}, and projection-based -- also known as ``safety filter'' -- RL methods for maintaining hard safety constraints, typically achieved through the use of control barrier functions (CBFs) \citep{cheng2019end}. Though CMDP-based methods enjoy convergence guarantees, they encourage safety without guaranteeing it, and cannot provide guarantees for hard safety constraints commonly required in physical systems. Likewise, methods like \cite{berkenkamp2017safe} do better by offering high-probability safety assurances, but stop short of guaranteeing safety. In systems where safety is critical, methods like \cite{cheng2019end} that provably guarantee hard constraint satisfaction are necessary.
However, the interaction between imposition of hard constraints and optimality of the resulting control policies is a subtle issue in RL. While projection-based safety-filter approaches \citep{wabersich2023data} provably guarantee safety, the projection procedure undermines any convergence guarantees enjoyed by the underlying RL methods.

In this work, we develop a class of model-free policy gradient methods that maintain safety or other stability properties by sampling directly from the set of state-dependent safe actions. The key to our approach is that we consider \textit{truncated} versions of commonly used stochastic policies, allowing us to sample directly from the safe action set at each state. This allows us to recover convergence guarantees by extending existing results for policy gradient methods to truncated policies. Our approach is applicable to a wide class of safety constraints including control barrier functions (CBFs), that enforce forward invariance of a set characterized by nonlinearly coupled states and actions \citep{ames2019control,ames2016control}, and reachability-type constraints \citep{wabersich2023data}. 
In addition to our theoretical results, we experimentally validate the practical utility of sampling-based safety-preservation methods by considering a special case: Beta policies with state-dependent, control barrier function (CBF)-constrained action sets. This novel approach extends the Beta policies in \cite{chou2017improving} to the state-dependent action constraint setting. Finally, we train the resulting CBF-constrained Beta policies using PPO to solve a safety-constrained inverted pendulum problem as well as a quadcopter navigation and obstacle avoidance problem, and compare the latter to a safety filter-based benchmark.\footnote{Our implementation is publicly available at \url{https://github.com/sharma1256/cbf-constrained_ppo}.} These case studies illustrate that our method simultaneously guarantees safety throughout training and guarantees optimality, even where existing benchmarks fail. 

\subsection{Related Work}
%
Safety and stability have seen a great deal of interest in recent years at the intersection of the RL and control communities (see \cite{brunke2022safe,garcia2015comprehensive} for overviews). We are interested in safety definitions that impose \textit{hard constraints} on the states and control actions (rather than, e.g., those used in robust RL~\citep{wiesemann2013robust, aswani2013provably} or RL for CMDPs \citep{achiam2017constrained, paternain2019constrained, ma2021model, bai2022achieving}). Model-based methods for guaranteeing stability using RL controllers in systems with known or learnable dynamics have been developed in \cite{berkenkamp2017safe,fazel2018global,zhang2021policy}. Recently, techniques leveraging control barrier functions to maintain safety \citep{cheng2019end} and dissipativity \citep{kosaraju2020reinforcement} have been developed. 

Our work lies in the model-free RL setting. The two dominant approaches in model-free RL are value function and policy gradient-based methods \citep{sutton2018reinforcement}. We focus on the latter in this paper. Since their origins early in the development of RL \citep{sutton2000policy, borkar2005actor, bhatnagar2009natural}, policy gradient methods have become the model-free algorithms of choice for complex problems with continuous, high-dimensional state and action spaces \citep{lillicrap2015continuous, schulman2017proximal, haarnoja2018soft}. Recent works have improved our understanding of gradient estimation procedures, global optimality properties, and convergence rates of these algorithms \citep{agarwal2020optimality, zhang2020global, suttle2023beyond}.
Popular approaches for safety in model-free RL include using bounds resulting from Gaussian process models~\citep{schreiter2015safe,rasmussen2003gaussian,sui2015safe}, reward-shaping, constrained policy optimization~\citep{achiam2017constrained,wachi2018safe}, and teacher advice~\citep{abbeel2004apprenticeship}. Our work is most closely related to those approaches that use a hard safe set specification and constraints on control inputs, e.g., control barrier functions~\citep{cheng2019end,fisac2018general,li2018safe,kosaraju2020reinforcement}. In particular, our key contribution is a model-free safe RL algorithm with convergence guarantees and provable safety guarantees under hard constraints like CBFs, even during training.

%% file: AISTATS2024/sections/Problem_Setting.tex
\section{PROBLEM SETTING} \label{sec:problem_setting}

Consider a discounted MDP $(\mathcal{X}, \mathcal{U}, \mathcal{P}, r, \gamma)$, where $\mathcal{X} \subseteq \mathbb{R}^m$ is the state space, $\mathcal{U} \subseteq \mathbb{R}^n$ is the action space, $\mathcal{P}(\cdot | x, u)$ is the transition probability function given action $u$ is taken in state $x$, $r : \mathcal{X} \times \mathcal{U} \rightarrow \mathbb{R}$ is the reward function, and $\gamma \in [0, 1]$ is the discount factor.
%
%
The MDP, which can be used to model a wide array of discrete-time systems, proceeds as follows: at time $k$, the system is in state $x_k$; a control input $u_k$ is applied to the system; a reward $r(x_k, u_k)$ is received; the system transitions into state $x_{k+1}$ according to the distribution $P(\cdot | x_k, u_k)$. The goal in this problem formulation is to maximize the expected discounted reward, which we define in \eqref{eqn:discounted_obj_fn} below. Note that deterministic dynamics can be recovered by imposing that, for each $x \in \mathcal{X}, u \in \mathcal{U}$, there exists $x' \in \mathcal{X}$ such that $\mathcal{P}(x' | x, u) = 1$. This is useful for modeling discretizations of continuous-time control problems, for example.
We assume throughout this paper that the dynamics are deterministic in this way, which is a common setting in safe control problems. Let $\mathcal{T} : \mathcal{X} \times \mathcal{U} \rightarrow \mathcal{X}$ represent the dynamics of the MDP, i.e., given state $x$ and control input $u$, $\mathcal{T}(x,u)$ denotes the state the system transitions into when input $u$ is applied while in state $x$.


Letting $\Delta(\mathcal{U})$ denote the set of all probability distributions over the set $\mathcal{U}$, a stochastic policy $\pi : \mathcal{X} \rightarrow \Delta(\mathcal{U})$ is a function mapping states to probability distributions over the action space $\mathcal{U}$. In other words, given a state $x$, an agent using policy $\pi$ will choose a control action $u \in \mathcal{U}$ by sampling $u \sim \pi(\cdot | s)$. For our purposes it will be useful to consider policies $\pi_{\theta}$ parameterized by $\theta \in \Theta \subseteq \mathbb{R}^k$, for some $k \ll |\mathcal{X}| \cdot |\mathcal{U}| = m \cdot n$, where $\Theta$ is a compact set of permissible parameters.

%
Let $\mathcal{S} \subset \mathcal{X}$ denote some ``safe'' or stable set within which we wish to keep the system. Furthermore, let $\mathbb{P}(S)$ denote the powerset of a set $S$, and consider a set-valued function $C : \mathcal{X} \rightarrow \mathbb{P}(\mathcal{U})$ given by
%
%
\begin{equation}
    C(x) = \{ u \in \mathcal{U} \ | \ \mathcal{T}(x, u) \in \mathcal{S} \}. \label{eqn:cbf_set}
\end{equation}
Intuitively, $C(x)$ is the set of all control inputs which when applied at state $x$ keep the system within the safe set at the next time step. We assume throughout that, for a given $x \in \mc{X}$, $C(x)$ is known. \wes{Since our primary focus is resolving the open problem of simultaneously guaranteeing convergence and hard safety constraint satisfaction, we leave the issue of learning or approximating $C(x)$ while maintaining these guarantees to future work.}
The general formulation can be used to accommodate a variety of notions of safety, including forward invariance, stability, and dissipativity enforced by, for example, CBFs and exponential CBFs (ECBFs), and control Lyapunov functions (CLFs) (see the supplementary material for an overview and \cite{ames2019control} for a comprehensive survey). As we will demonstrate in the case studies below, the use of our method in conjunction with (E)CBFs is particularly natural to provide guarantees in problems with hard safety constraints.
%
%
To ensure that we can sample from $C(x)$ and integrals over $C(x)$ are well-defined, we make the following assumption. Let $\mu$ denote the Lebesgue measure.
\begin{assumption} \label{assum:muC_bounded}
    There exist $m, M > 0$ such that $m \leq \mu(C(x)) \leq M$, for all $x \in \mathcal{X}$. Furthermore, $\cup_{x \in \mathcal{X}} C(x)$ is compact.
\end{assumption}

Given a policy $\pi_{\theta}$, consider the distribution $\pi_{\theta}^C( \cdot | x)$ obtained by truncating $\pi_{\theta}( \cdot | x)$ to the set $C(x)$. More precisely:
\begin{equation} \label{eqn:piC_definition}
    \pi^C_{\theta}(u | x) =
    \begin{cases}
        \frac{\pi_{\theta}(u | x)}{\pi_{\theta}(C(x) | x)} & u \in C(x) \\
        0 & u \notin C(x),
    \end{cases}
\end{equation}
where $\pi_{\theta}(C(x) | x) = \int_{C(x)} \pi_{\theta}(u | x) du$.
As long as we can check membership in $C(x)$ for any given $u \in \mathcal{U}$, and assuming that the volume of $C(x)$ is strictly positive, for all $x \in \mathcal{X}$, we can generate from this distribution by using rejection sampling, i.e. repeatedly sampling $u \sim \pi_{\theta}(\cdot | x)$ until we obtain $u \in C(x)$. Note that, depending on the structure of parametrized policies $\pi^C_{\theta}$, if $C(x)$ has a particularly nice form, such as an interval or hyperrectangle, there may be more efficient methods than rejection sampling for sampling from the truncated distribution $\pi^C_{\theta}(\cdot | x)$ directly. \wes{We exploit this fact when leveraging Beta policies in the experimental results of Section \ref{sec:experiments} below.}

With this setup in mind, and given a fixed start state $x_0$, we propose a policy gradient-based algorithm maximizing the objective function
\begin{equation} \label{eqn:discounted_obj_fn}
    J(\theta) = \mathbb{E}_{\pi_{\theta}^C} \left[ \sum_{k=0}^{\infty} \gamma^k r(x_k, u_k) \ \Big| \ x_0 \right],
\end{equation}
the expected discounted reward under policy $\pi^C_{\theta}$.
Before proceeding with describing and analyzing the algorithm, we first need to identify conditions that ensure that, for each policy parameter $\theta$, taking expectations with respect to $\pi_{\theta}^C$ is well-defined and thus meaningful. In order for \eqref{eqn:discounted_obj_fn} to be well-defined, we need to know that, for each policy parameter $\theta$, the occupancy measure of the Markov chain induced by $\pi_{\theta}^C$ on $\mathcal{S}$ is irreducible and satisfies certain ergodicity conditions. Once these are proven, we will be justified in performing gradient ascent on the objective function \eqref{eqn:discounted_obj_fn}.

%% file: AISTATS2024/sections/Theoretical_Results.tex
\section{THEORETICAL RESULTS}

In this section we develop the theory underlying our sampling-based method for \wes{RL with hard safety constraints.}
Our key contributions include proving that \eqref{eqn:discounted_obj_fn} is well-defined (\S\ref{subsec:obj_well_defined}), obtaining gradient expressions for it from which we can sample (\S\ref{subsec:policy_gradients}), and developing and establishing the convergence of a policy gradient algorithm for optimizing \eqref{eqn:discounted_obj_fn} (\S\ref{subsec:algorithm} and \S\ref{subsec:convergence}).
%
%
All proofs are deferred to the supplementary material.
%
\wes{It is important to note that, though we assumed the deterministic dynamics common to safe control in \S\ref{sec:problem_setting}, all our theoretical results go through in the stochastic dynamics case under standard ergodicity assumptions. Our key theoretical contribution in what follows is to show that, even in the \textit{deterministic} dynamics case, we can ensure that the objective is well-defined (\S\ref{subsec:obj_well_defined}) and obtain convergence (\S\ref{subsec:convergence}).
%
}

\subsection{Discounted Return is Well-defined} \label{subsec:obj_well_defined}

First, we show that the objective \eqref{eqn:discounted_obj_fn} is well-defined when using truncated policies, even in continuous spaces systems with deterministic dynamics.
Our key contribution in this setting is to ensure that, given reasonable conditions on the policies under consideration, important ergodicity properties of their induced Markov chains hold. This fact, established in Proposition \ref{prop:irreducibility} and Corollary \ref{cor:irreducibility}, is nontrivial and its proof relies on a careful analysis of the propagation of probability mass through the transition dynamics and an interesting application of the Lebesgue-Radon-Nikodym Theorem \cite[\S3.2]{folland1999real}.
As in the previous section, let $\mathcal{S} \subset \mathbb{R}^m$ denote the ``safe'' set within which the system remains so long as all control inputs are selected from $C(x)$. Though we leave open the possibility that $\mathcal{S}$ satisfies a more specific stability conditions rather than a generic notion of ``safety'', we will typically use the term ``safe'' for ease of presentation. Let $\mu$ denote Lebesgue measure. We make the following definition:
\begin{definition}
The Markov chain $\{x_k\}_{k \in \mathbb{N}}$ induced by $\pi^C_{\theta}$ on $\mathcal{X}$ is $\mu$-irreducible on $\mathcal{S}$ if, for any $\mu$-measurable $\mathcal{B} \subset \mathcal{S}$, if $\mu(\mathcal{B}) > 0$, then $\sum_{k \in \mathbb{N}} P(x_k \in \mathcal{B} \ | \ x_0 = x) > 0$, for all $x \in \mathcal{S}$.
\end{definition}

This means that, for a Markov chain to be ($\mu$-)irreducible on the safety set, all safe subsets with positive volume must be reachable from any initial safe state with positive probability. Notice that $\{ x_k \}_{k \in \mathbb{N}}$ is in fact a Markov chain on the safe set $\mathcal{S}$ since, by the definition of $C(x)$, only those control inputs keeping the system within $\mathcal{S}$ are allowed.
In the sequel, we will prove that, for each $\theta$, under suitable conditions the Markov chain induced by $\pi_{\theta}^C$ on $\mathcal{S}$ is irreducible and the objective \eqref{eqn:discounted_obj_fn} is thus well-defined, which is a prerequisite for developing policy gradient methods based on it. See \cite[\S2.3]{konda2002thesis} for details on irreducibility in this setting.


Given an element $x \in \mathcal{S}$ and dynamics $\mathcal{T}$, let $R(x) \subset \mathcal{S}$ consisting of all elements reachable in one step from $x$ under $\mathcal{T}$. Furthermore, for $\mathcal{A} \subset \mathcal{S}$, define $R(\mathcal{A}) = \cup_{x \in \mathcal{A}} R(x)$. Also, given $\varepsilon > 0$ and $x \in \mathbb{R}^m$, let $B_{\varepsilon}(x)$ denote the open ball of radius $\varepsilon$ centered at $x$. Finally, for $\mathcal{A} \subset \mathcal{S}$, define
%
%
$\mathcal{T}^{-1}_x (\mathcal{A}) := \{ u \in \mathcal{U} \ | \ T(x, u) \in \mathcal{A} \}.$
Intuitively, $\mathcal{T}^{-1}_x(\mathcal{A})$ is the set of all control inputs that, when taken in state $x$, drive the system into $\mathcal{A}$.
The following assumptions are needed in what follows.
\begin{assumption} \label{assum:volume_preservation}
    For any $x \in \mathcal{S}$ and any $\mu$-measurable set $\mathcal{A} \subset R(x)$, $\mu(\mathcal{A}) > 0$ if and only if $\mu(\mathcal{T}^{-1}_x(\mathcal{A})) > 0$.
\end{assumption}

Assumption \ref{assum:volume_preservation} ensures the system dynamics map positive volume subsets of control inputs to positive volume subsets of the state space and vice versa, which is important for our application of the Lebesgue-Radon-Nikodym Theorem in Proposition \ref{prop:irreducibility}. It is satisfied by systems where control inputs have a measurable effect on each entry in the next state vector and thus encompasses a wide array of potentially nonlinear systems. 
%
%
\begin{assumption} \label{assum:positive_probability}
    For any $\theta \in \Theta$, where $\Theta$ is the set of permissible policy parameters, for any element in the safe set $x \in \mathcal{S}$, and for any set $\mathcal{A} \subset C(x)$ satisfying $\mu(\mathcal{A}) > 0$, the policy $\pi^C_{\theta}(\cdot | x)$ assigns positive probability to $\mathcal{A}$, i.e. $\int_\mathcal{A} \pi^C_{\theta}(a | x) da > 0$.
\end{assumption}

Assumption \ref{assum:positive_probability}, which is standard in the RL literature, ensures that any set of allowable control inputs that has strictly positive volume will be sampled from with strictly positive probability.
\begin{assumption} \label{assum:positive_volume_reachability}
    For each $x \in \mathcal{S}$, $\mu(R(x)) > 0$, and, given $\mathcal{B} \subset \mathcal{S}$, there exists $n \in \mathbb{N}$ such that $\mathcal{B}$ is reachable in $n$ steps from $x$.
\end{assumption}

The conditions imposed in Assumption \ref{assum:positive_volume_reachability} guarantee that, for any state $x \in \mathbb{X}$: (i) the set of states reachable from $x$ in one step has strictly positive volume; (ii) any subset of the safe set $\mathcal{S}$ is reachable in at most $n$ steps from $x$. These conditions are closely related to the familiar notion of controllability of control theory.
Under these conditions, we have the following proposition and its immediate corollary.


\begin{proposition} \label{prop:irreducibility}
    Under Assumptions \ref{assum:volume_preservation}, \ref{assum:positive_probability}, \ref{assum:positive_volume_reachability}, for given $\theta$ and any subset $\mathcal{B} \subset \mathcal{S}$ satisfying $\mu(\mathcal{B}) > 0$, the Markov chain induced by $\pi^C_{\theta}$ on $\mathcal{S}$ enters $\mathcal{B}$ with strictly positive probability.
\end{proposition}

\begin{corollary} \label{cor:irreducibility}
    $\{ x_n \}$ is $\mu$-irreducible on $\mathcal{S}$.
\end{corollary}

Now that we are assured that the objective function \eqref{eqn:discounted_obj_fn} is well-defined, we are justified in attempting to perform gradient ascent on it. In order to accomplish this, however, we need access to gradient estimates. This is the subject of the next section.

\subsection{Policy Gradients} \label{subsec:policy_gradients}

Despite the presence of $C$ in $\pi_{\theta}^C$, under mild assumptions on the underlying policy $\pi_{\theta}$, we can apply the classic policy gradient theorem of \cite{sutton2000policy} to \eqref{eqn:discounted_obj_fn} to obtain a gradient expression from which we can sample. Let $d_{\theta}^C(\cdot) := (1 - \gamma) \sum_{k=0}^{\infty} \gamma^t P(x_k \in \cdot \ | \ \pi_{\theta}^C)$ denote the discounted state occupancy measure of the Markov chain induced by policy $\pi_{\theta}^C$ on $\mathcal{S}$. Furthermore, let
$Q^{\pi_{\theta}^C}(x, u) = \mathbb{E}_{\pi_{\theta}^C} \left[ \sum_{k=0}^{\infty} \gamma^k r(x_k, u_k) \ | \ x_0 = x, u_0 = u \right].$
We make the following assumption:
\begin{assumption} \label{assum:differentiability}
    $\pi_{\theta}(u | x) > 0$ and $\pi_{\theta}(u | x)$ is differentiable in $\theta$, for all $x \in \mathcal{X}, u \in \mathcal{U}$.
\end{assumption}
Recall from \eqref{eqn:piC_definition} that $\pi^C_{\theta}( \cdot | x)$ is simply the probability density function $\pi_{\theta}(\cdot | x)$ truncated to the set $C(x)$.
Note, since the value of $C(x)$ at a given $x$ is independent of $\theta$, we can take the derivative inside the integral sign in the latter expression to obtain $\nabla \pi_{\theta}(C(x) | x) = \int_{C(x)} \nabla \pi_{\theta}(u | x) du$, so $\pi^C_{\theta}(C(x) | x)$ is differentiable.
Given these facts, combined with Assumption \ref{assum:differentiability}, the above expression for $\pi^C_{\theta}$ implies that, for any $x \in \mathcal{S}$, the policy $\pi^C_{\theta}(u | x)$ is differentiable with respect to $\theta$, for any $u \in C(x)$. \wes{In short, $\pi^C_{\theta}$ satisfies its own version of Assumption \ref{assum:differentiability}, which we formalize in the following:
\begin{lemma} \label{lemma:piC_differentiability}
    $\pi^C_{\theta}(u | x) > 0$ and $\pi^C_{\theta}(u | x)$ is differentiable in $\theta$, for all $x \in \mathcal{S}$ and $u \in C(x)$.
\end{lemma}
}

The policy gradient theorem \citep{konda2002thesis} implies
%
\begin{equation} \label{eqn:pgt}
    %
    \nabla J(\theta) = \frac{1}{1 - \gamma} \mathbb{E}_{\pi^C_{\theta}} \left[ Q^{\pi_{\theta}^C}(x, u) \nabla \log \pi_{\theta}^C(u | x)  \right].
\end{equation}

In order to carry out gradient updates based on this expression, we first need to be able to estimate $\nabla_{\theta} \log \pi^{C}_{\theta}(u | x) = \nabla_{\theta} \pi^C_{\theta}(u | x) / \pi^C_{\theta}(u | x)$, for arbitrary $u, x$. We will discuss how to estimate $Q^{\pi^C_{\theta}}(x, u)$ in an unbiased manner in the following section. Since we already have access to $\pi^C_{\theta}(u | x)$, we can focus on estimating $\nabla_{\theta} \pi^C_{\theta}(u | x)$.
Based on \eqref{eqn:piC_definition}, the gradient of $\pi^C_{\theta}(u | x)$ with respect to $\theta$ is
\begin{align}
    \nabla &\pi^C_{\theta}(u | x) = \nabla \left[ \frac{\pi_{\theta}(u | x)}{\pi_{\theta}(C(x) | x)} \right] \\
    &= \frac{\nabla \pi_{\theta}(u | x)}{\pi_{\theta}(C(x) | x)} - \frac{\pi_{\theta}(u | x)}{[\pi_{\theta}(C(x) | x)]^2} \nabla \pi_{\theta}(C(x) | x) \\
    &= \frac{1}{\pi_{\theta}(C(x) | x)} \left[ \nabla \pi_{\theta}(u | x) - \pi_{\theta}(u | x) \nabla \log \pi_{\theta}(C(x) | x) \right].
\end{align}
To estimate $\pi_{\theta}(C(x) | x)$, we need to be able to estimate $\int_{C(x)} \pi_{\theta}(u | x) du$. Given access to $\pi_{\theta}(\cdot | x)$ and $C(x)$, we can use numerical integration or Monte Carlo techniques to approximate this integral. The standard Monte Carlo approach is to uniformly sample $M$ elements $u_i \sim U(C(x))$ from $C(x)$, then estimate
\begin{equation} \label{eqn:pi_estimate}
    \widehat{\pi_{\theta}}(C(x) | x) = \mu(C(x)) \frac{1}{M} \sum_{i=1}^M \pi_{\theta}(u_i | x),
\end{equation}
where $\mu(C(x))$ is the volume of $C(x)$. This estimate is based on the fact that
\begin{align}
    \small
    \pi_{\theta}&(C(x) | x) = \int_{C(x)} \pi_{\theta}(u | x) du \\
    &= \mu(C(x)) \int_{C(x)} \frac{\pi_{\theta}(u | x)}{\mu(C(x))} du \\
    &= \mu(C(x)) E_{u \sim U(C(x))} [\pi_{\theta}(u | x)] \\
    &= \mu(C(x)) \lim_{M \rightarrow \infty} \frac{1}{M} \sum_{i=1}^M \pi_{\theta}(u_i | x),
    \normalsize
\end{align}
where the last equality holds by the law of large numbers. Since $C(x)$ is fixed given $x$, gradient estimates $\widehat{\nabla \log \pi_{\theta}}(C(x) | x)$ and ultimately $\widehat{\nabla \log \pi^C_{\theta}}(u | x)$ can also be obtained by estimating the integral $\int_{C(x)} \nabla \pi_{\theta}(u | x) du$. In the Monte Carlo situation, this can be obtained from \eqref{eqn:pi_estimate} by differentiating each term with respect to $\theta$.

\subsection{Algorithm} \label{subsec:algorithm}
In this section, we present a hard safety-constrained random-horizon policy gradient (Safe-RPG) algorithm. Our algorithm is based on the random-horizon policy gradient (RPG) scheme developed in \cite{zhang2020global}, which uses a random rollout horizon and recent advances in non-convex optimization to obtain unbiased policy gradient estimates and ensure finite-time convergence to approximately locally optimal policies. As discussed in the following section, our convergence results ensure asymptotic convergence of Algorithm \ref{alg:cbf-rpg} to a stationary point of \eqref{eqn:discounted_obj_fn}, but can likely be strengthened to prove finite-time convergence to approximately locally optimal policies. The main algorithm is presented in Algorithm \ref{alg:cbf-rpg}, which depends on the action-value function estimation subroutine in Algorithm \ref{alg:estq}.

\subsection{Convergence} \label{subsec:convergence}
In this section we
%
%
show asymptotic convergence of Algorithm \ref{alg:cbf-rpg} to the set of stationary points of \eqref{eqn:discounted_obj_fn}.
The key challenge in this result revolves around the need to establish that the policies we consider satisfy important differentiability and continuity properties, which necessitates a careful analysis of the Lipschitz properties of the score functions of our truncated policies in the proof of Lemma \ref{lemma:piC_assumptions_hold}. To proceed, we need the following assumption on the reward function $r$ and underlying, untruncated policy class $\{ \pi_{\theta} \}_{\theta \in \Theta}$.

\begin{assumption} \label{assum:r_and_pi}
    The reward function $r$ and parameterized policy class $\{ \pi_{\theta} \}_{\theta \in \Theta}$ satisfy the following:
    \begin{enumerate}
        \item The absolute value of the reward $r$ is uniformly bounded, i.e., there exists $U_r$ such that $0 \leq \sup_{(x, u) \in \mathcal{X} \times \mathcal{U} } | r(x, u) | \leq U_r$.
        \item 
        For all $x \in \mathcal{X}, u \in \mathcal{U}$, $\nabla \log \pi_{\theta}(u | x)$ exists, and there exist $L_{\Theta} \geq 0$ and $B_{\Theta} \geq 0$ such that, for all $x \in \mathcal{X}, u \in \mathcal{U}$,
        \begin{enumerate}
            \item $\norm{ \nabla \log \pi_{\theta}(u | x) - \nabla \log \pi_{\theta'}(u | x) } {\leq}L_{\Theta} \norm{\theta - \theta'}$, for all $\theta, \theta' \in \Theta$
            \item $\norm{\nabla \log \pi_{\theta}(u | x)} \leq B_{\Theta}$, for all $\theta \in \Theta$.
        \end{enumerate}
    \end{enumerate}
\end{assumption}
%
%
    
    
%
%
\begin{algorithm}[h]
    \SetAlgoLined
    \caption{ \texttt{EstQ:} Unbiasedly Estimating $Q$} \label{alg:estq}
    \KwData{$x, u, \theta$.}
    \KwResult{Unbiased estimate of $Q^{\pi^C_{\theta}}(x, u)$.}
    \BlankLine
    {\bf Initialization:} Sample $T \sim \text{Geom}(1 - \gamma^{1/2})$ and initialize $\hat{Q} \gets 0, x_0 \gets x, u_0 \gets u$. \\
    \For{$t = 0, \ldots, T - 1$}{
        $\hat{Q} \gets \hat{Q} + \gamma^{t / 2} r(x_t, u_t)$ \\
        $x_{t+1} \sim \mathcal{P}(\cdot | x_t, u_t)$ \\
        $u_{t+1} \sim \pi^C_{\theta_k}(\cdot | x_{t+1})$
    }
    $\hat{Q} \gets \hat{Q} + \gamma^{T/2} r(x_T, u_T)$ \\
    \Return{$\hat{Q}$}
\end{algorithm}
\begin{algorithm}[h]
    \SetAlgoLined
    \caption{ \texttt{Safe-RPG:} Hard safety-constrained Random-horizon Policy Gradient} \label{alg:cbf-rpg}
    \KwData{$x_0, \theta_0$, Monte Carlo sample size $M$.}
    \KwResult{Locally optimal policy}
    \BlankLine
    {\bf Initialization:} Set $k \gets 0$. \\
    \Repeat{convergence}{
        Sample $T_{k+1} \sim \text{Geom}(1 - \gamma)$, $u_0 \sim \pi^C_{\theta_k}(\cdot | x_0)$. \\
        \For{$t = 0, \ldots, T_{k+1} - 1$}{
            $x_{t+1} \sim \mathcal{P}(\cdot | x_t, u_t)$ \\
            $u_{t+1} \sim \pi^C_{\theta_k}(\cdot | x_{t+1})$
        }
        $\hat{Q}^{\pi^C_{\theta_k}}(x_{T_{k+1}}, u_{T_{k+1}}) = \texttt{EstQ}(x_{T_{k+1}}, u_{T_{k+1}}, \theta_k)$ \\
        Uniformly sample $\{ u_l \}_{l=1, \ldots, M}$ from $C(x_{T_{k+1}})$, then use them to compute $\widehat{\nabla \log \pi^C_{\theta_k}}(u_{T_{k+1}} | x_{T_{k+1}})$ \\
        \normalsize
        $\theta_{k+1} \gets \theta_k + \frac{ \alpha_k }{ 1 - \gamma } \hat{Q}^{\pi^C_{\theta_k}} ( x_{T_{k+1}}, u_{T_{k+1}} ) \widehat{\nabla \log \pi^C_{\theta_k}}(u_{T_{k+1}} | x_{T_{k+1}})$ \\
        \normalsize
        $k \gets k + 1$ \\
    }
\end{algorithm}
%

%
%
Assumptions \ref{assum:differentiability} and \ref{assum:r_and_pi} were used to prove asymptotic convergence of the RPG algorithm with \textit{untruncated} policies to stationary points in \cite[Theorem~4.4]{zhang2020global}. For an analogous result to apply to the truncated policies we consider, it must be shown that the Lipschitz and differentiability conditions in part 2 of Assumption \ref{assum:r_and_pi} hold for the constrained policies $\{ \pi^C_{\theta} \}_{\theta \in \Theta}$. It turns out that, under the same conditions on the untruncated policy $\{ \pi_{\theta} \}_{\theta \in \Theta}$, these properties are automatically satisfied for $\{ \pi^C_{\theta} \}_{\theta \in \Theta}$.

\begin{lemma} \label{lemma:piC_assumptions_hold}
Under Assumptions \ref{assum:muC_bounded}, \ref{assum:differentiability}, and \ref{assum:r_and_pi}, $\nabla \log \pi^C_{\theta}(u | x)$ exists, for all $x \in \mathcal{X}, u \in \mathcal{U}$. Furthermore, there exist constants $L^C_{\Theta} \geq0$ and $B^C_{\Theta} \geq 0$ such that, for all $x \in \mathcal{X}, u \in \mathcal{U}$, \\
%
(i) $\norm{ \nabla \log \pi^C_{\theta}(u | x) - \nabla \log \pi^C_{\theta'}(u | x) } \leq L^C_{\Theta} \norm{\theta - \theta'}$, for all $\theta, \theta' \in \Theta$, and \\
    %
   (ii) $\norm{\nabla \log \pi^C_{\theta}(u | x)} \leq B^C_{\Theta}$, for all $\theta \in \Theta$.
%
\end{lemma}

With Lemma \ref{lemma:piC_assumptions_hold} in hand, we have the following result.
\begin{theorem} \label{thm:as_convergence}
Let Assumptions \ref{assum:positive_probability}, \ref{assum:positive_volume_reachability}, \ref{assum:differentiability}, and \ref{assum:r_and_pi} hold. Let $\{ \theta_k \}_{k \in \mathbb{N}}$ be the sequence generated by Algorithm \ref{alg:cbf-rpg} with stepsize sequence $\{ \alpha_k \}_{k \in \mathbb{N}}$ satisfying $\sum_{k=0}^{\infty} \alpha_k = \infty$ and $\sum_{k=0}^{\infty} \alpha_k^2 < \infty$. Then $\lim_k \theta_k \in \Theta^*$, where $\Theta^*$ is the set of stationary points of \eqref{eqn:discounted_obj_fn}.
\end{theorem}

\begin{remark} \label{rmk:2}
    \wes{
    By arguments analogous to those in the proof of \cite[Thm. 3]{zhang2019convergence}, it can also be shown that, under the same assumptions as in Theorem \ref{thm:as_convergence} and appropriate stepsize selection, Algorithm \ref{alg:cbf-rpg} achieves $\varepsilon$-approximate first-order stationarity with a finite-time sample complexity of $\mathcal{O}(\varepsilon^{-2})$.
    }
\end{remark}

Given Lemma \ref{lemma:piC_assumptions_hold}, the proof of the theorem follows directly from that of \cite[Theorem~4.4]{zhang2020global}. With suitable modifications to Algorithm \ref{alg:cbf-rpg} incorporating periodically increasing stepsizes, these results can likely be strengthened to obtain finite-time convergence to an $\varepsilon$-locally optimal policy using the machinery developed in \cite{zhang2020global}. We leave this to future work.

%% file: AISTATS2024/sections/Experimental_Results.tex
\section{EXPERIMENTAL RESULTS} \label{sec:experiments}

We now experimentally demonstrate the effectiveness of our sampling-based safe RL approach. Specifically, we evaluate the use of CBF-constrained Beta policies combined with the popular Proximal Policy Optimization (PPO) \citep{schulman2017proximal} algorithm on safety-constrained inverted pendulum and quadcopter navigation environments. The use of Beta policies with variable action space constraints allows us to directly sample from a CBF-constrained action space at each timestep. In addition to providing a practical example of how truncated policies can be used to ensure safety, this method extends the work \cite{chou2017improving} on the use of Beta policies for deep RL \siva{from constant to  state-dependent action space constraints}.
%

\begin{figure*}[htb]
     \centering
     \begin{subfigure}[t]{0.32\textwidth}
        \centering
        \includegraphics[width=1.1\linewidth, trim=0cm 0.6cm 0cm 0cm]{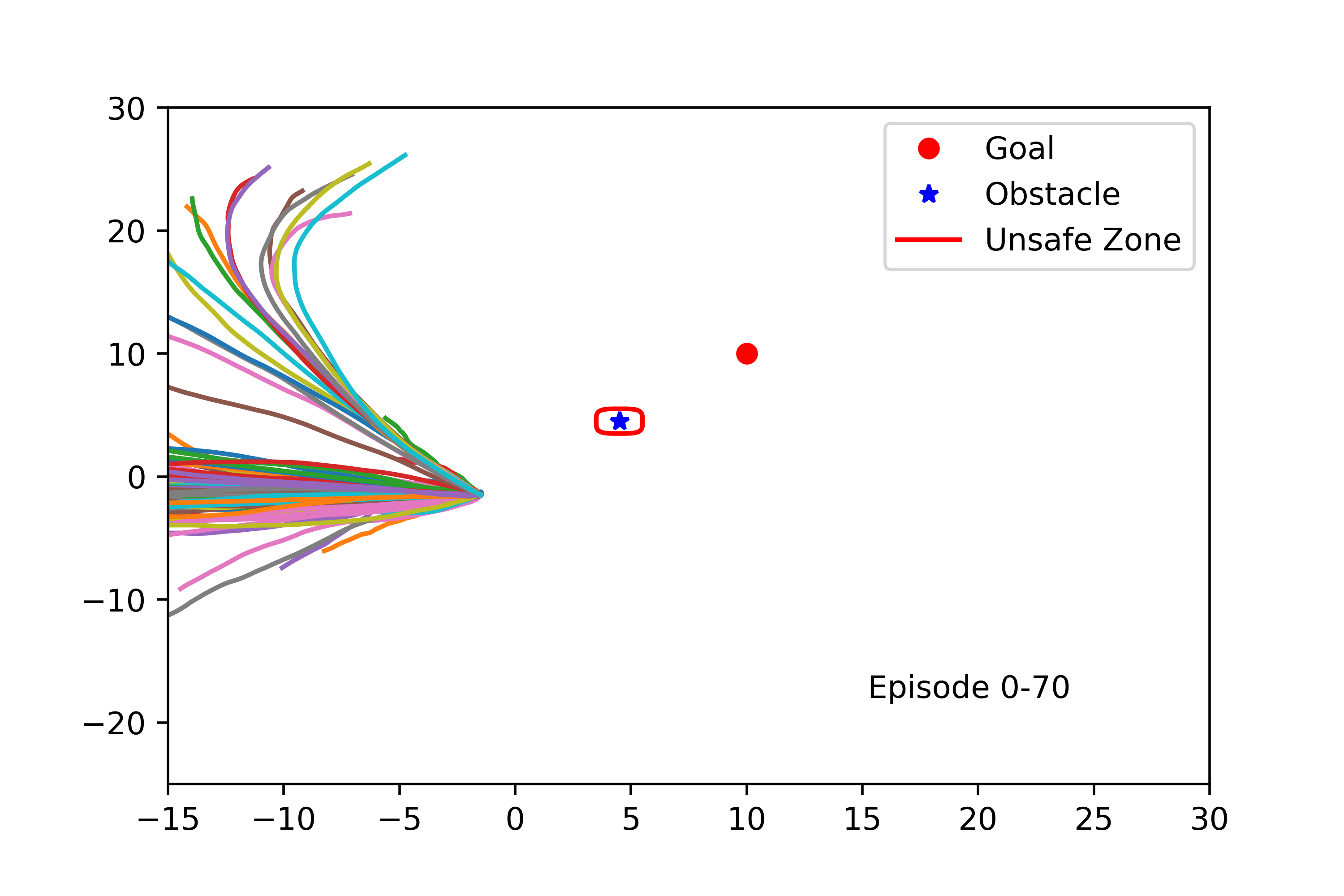}
        \caption{Initial exploration.}
        \label{fig:phase_1}
     \end{subfigure}
     \hfill
     \begin{subfigure}[t]{0.32\textwidth}
        \centering
        \includegraphics[width=1.1\linewidth, trim=0cm 0.6cm 0cm 0cm]{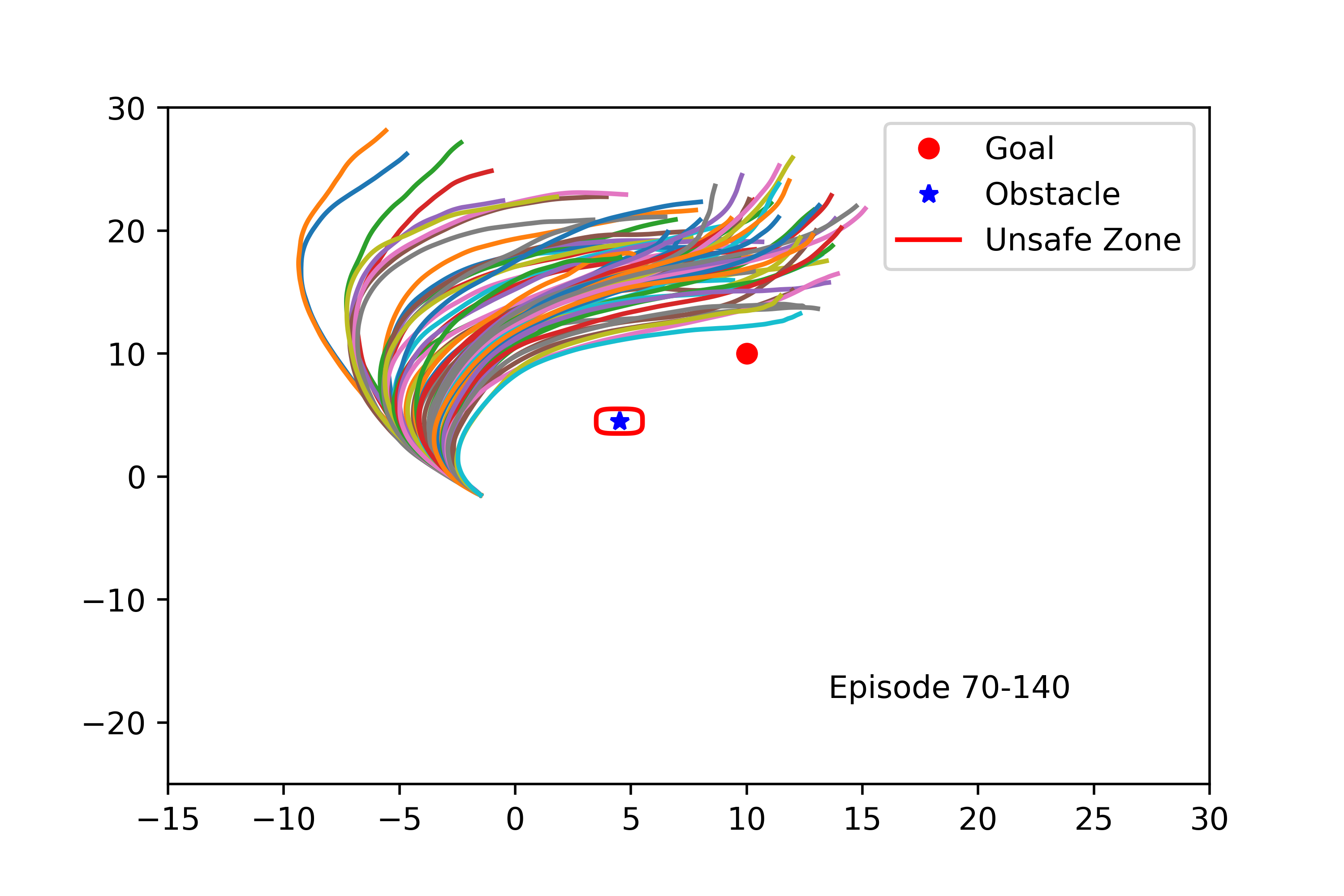}
        \caption{Discovering goal direction.}
        \label{fig:phase_2}
     \end{subfigure}
     \hfill
     \begin{subfigure}[t]{0.32\textwidth}
        \centering
        \includegraphics[width=1.1\linewidth, trim=0cm 0.6cm 0cm 0cm]{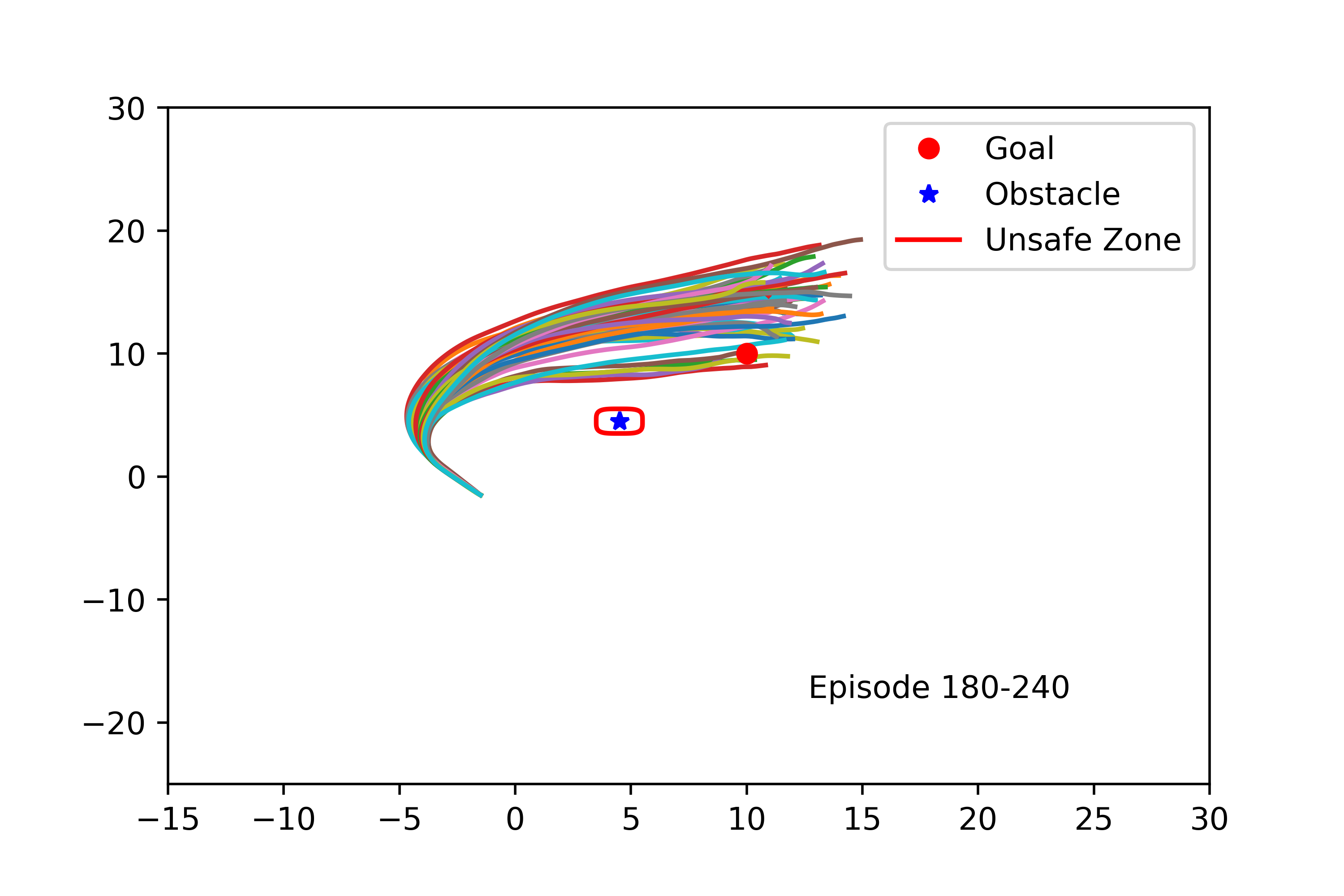}
        \caption{Reaching the goal.}
        \label{fig:phase_3}
     \end{subfigure}
     \vspace{-1mm}
        \caption{\textbf{Safety and convergence of CBF-Constrained Beta policies:} Agent was trained with PPO on the quadrotor navigation problem with an obstacle. Safety was maintained and goal was eventually reached.}
        \label{fig:quadcopter_experiments_beta}
    \vspace{2mm}
     \centering
     \begin{subfigure}[t]{0.32\textwidth}
        \centering
        \includegraphics[width=1.1\linewidth, trim=0cm 0.6cm 0cm 0cm]{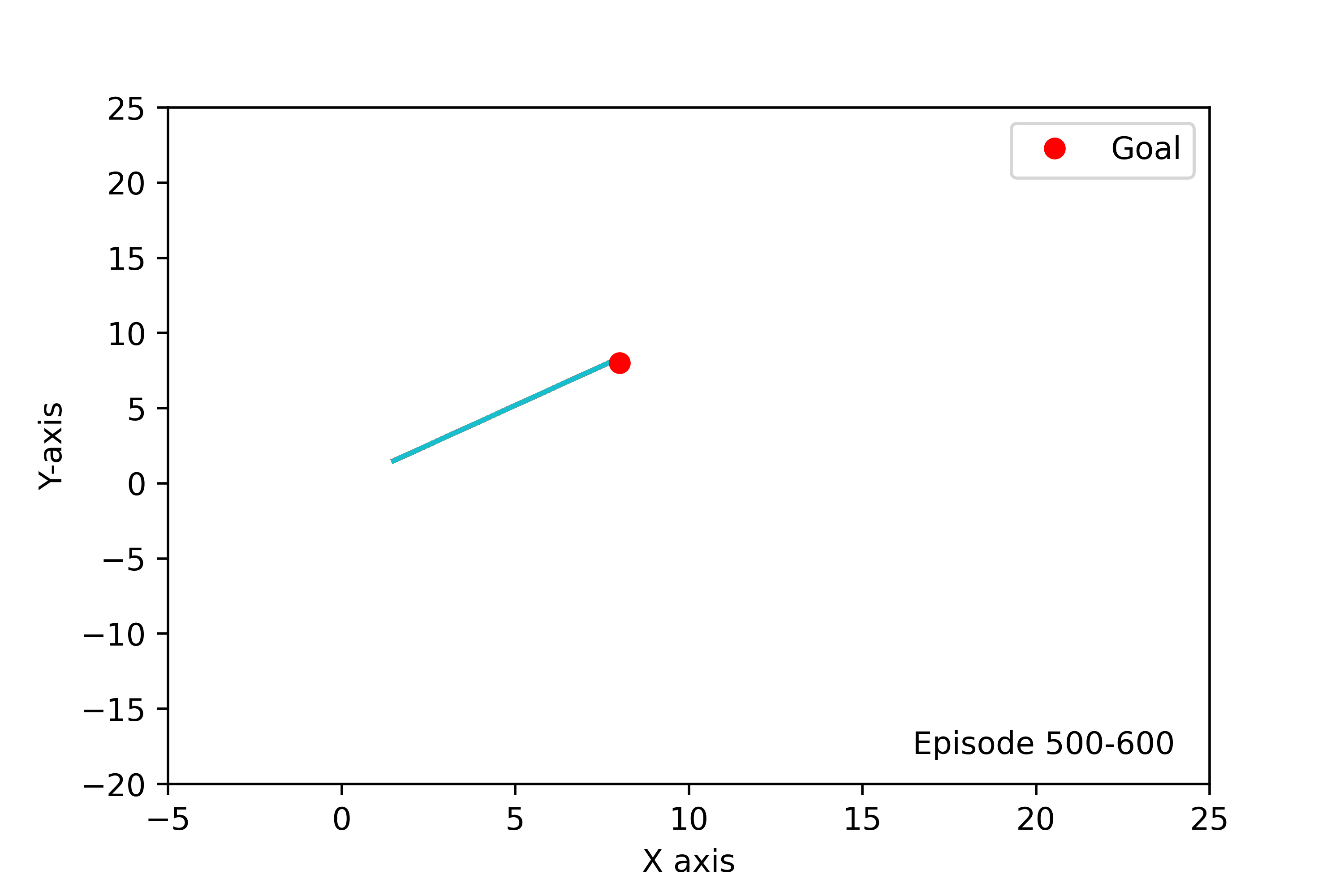}
        \caption{No obstacle.}
        \label{fig:phase_1_bench}
     \end{subfigure}
     \hfill
     \begin{subfigure}[t]{0.32\textwidth}
        \centering
        \includegraphics[width=1.1\linewidth, trim=0cm 0.6cm 0cm 0cm]{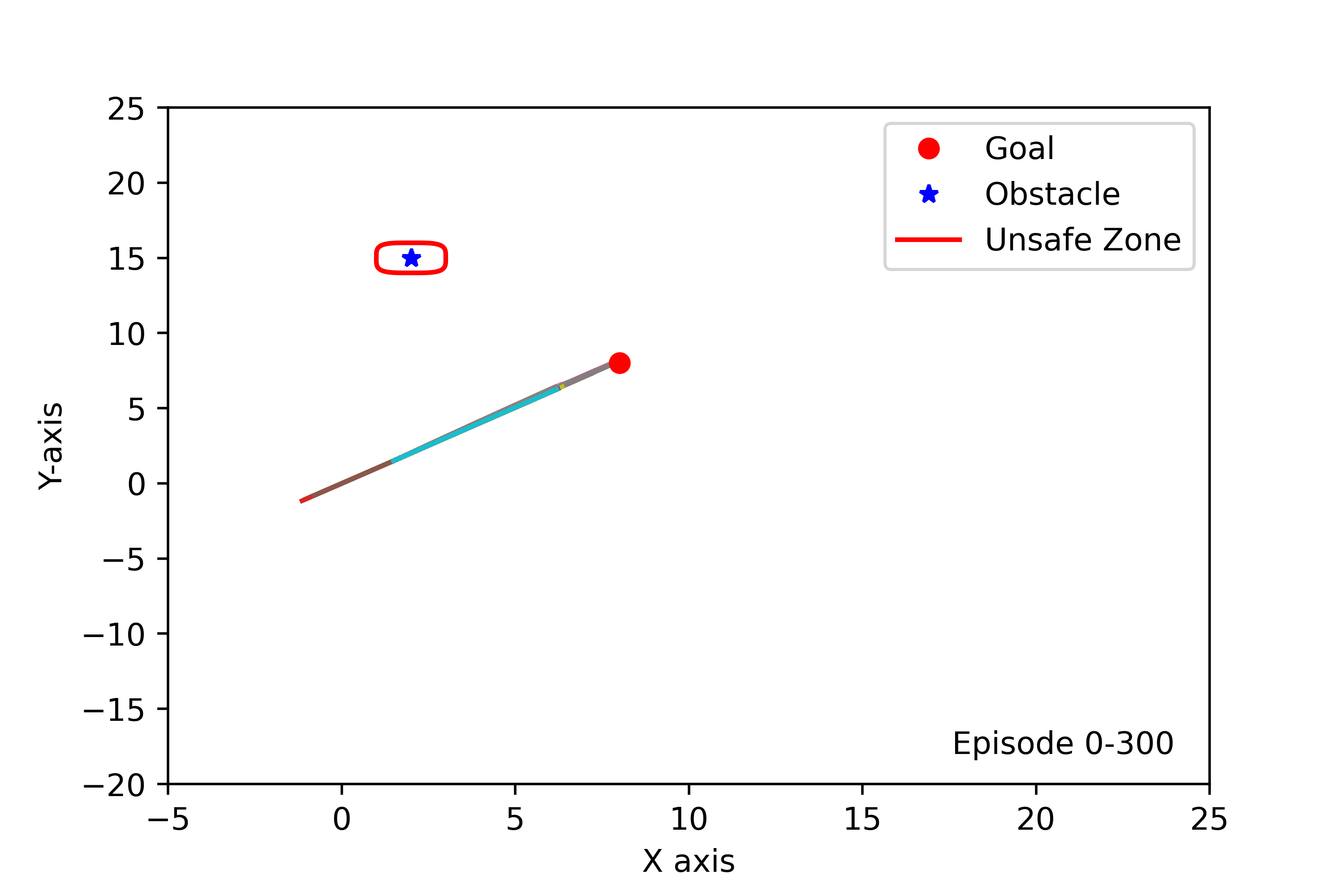}
        \caption{Distant obstacle.}
        \label{fig:phase_2_bench}
     \end{subfigure}
     \hfill
     \begin{subfigure}[t]{0.32\textwidth}
        \centering
        \includegraphics[width=1.1\linewidth, trim=0cm 0.6cm 0cm 0cm]{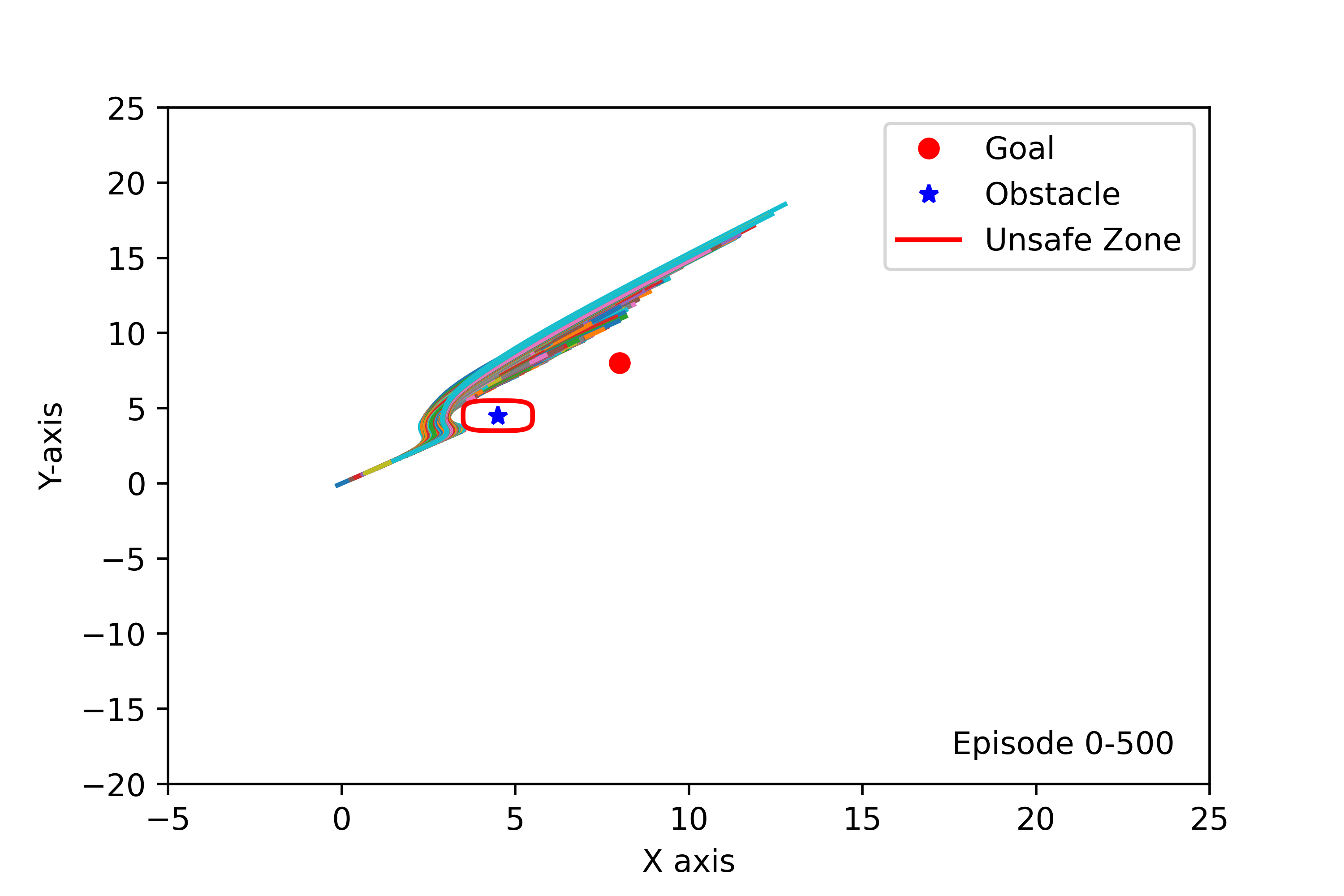}
        \caption{Interfering obstacle.}
        \label{fig:phase_3_bench}
     \end{subfigure}
       \vspace{-1mm}
        \caption{\textbf{Failure of benchmark safety-filtered Gaussian policies:} Agents were trained with PPO using three different obstacle configurations. When the obstacle is distant or non-existent, the method succeeds. When the obstacle is in the way, the resulting policy is suboptimal. In all cases, safety is maintained.}
        \vspace{-1em}
\label{fig:quadcopter_experiments_projection}
\end{figure*}

\subsection{CBF-Constrained Beta Policies}
When actions must be restricted to lie within fixed, predetermined bounds due to physical or numerical constraints, the common practice of simply clipping policies with infinite support (e.g., Gaussian policies) can cause bias and performance issues. \cite{chou2017improving} propose and leverage finite-support Beta distribution-based policies to overcome these issues. We extend this approach to obtain policies that sample directly from the safe control actions prescribed by the CBF at a given state.

%
%
%
In order to describe these CBF-constrained Beta policies, let us first recall the probability density function (p.d.f.) of a one-dimensional Beta distribution:
\vspace{-1mm}
\begin{equation}
    f(u; \alpha, \beta) = \frac{ \Gamma(\alpha + \beta) }{ \Gamma(\alpha) \Gamma(\beta) } u^{\alpha - 1} (1 - u)^{\beta - 1}, \label{eqn:beta_pdf}
\end{equation}
%
where $u \in [0, 1]$, $\alpha, \beta > 0$, and $\Gamma(z) = \int_0^{\infty} u^{z - 1} e^{-u} du$ is the Gamma function defined for $z \in \mathbb{C}$ with $\text{Re}(z) > 0$. A Beta policy sampling from the fixed interval $[0, 1]$ is given by $f(u; \alpha_{\theta}(x), \beta_{\theta}(x) )$, where $\alpha_{\theta}, \beta_{\theta} : \mathcal{X} \rightarrow \mathbb{R}^+$ are parameterized functions (e.g., neural networks) mapping states to the parameters $\alpha, \beta$ of the Beta distribution.
When the action space $\mc{U} \subset \mathbb{R}^n$ is of dimension $n \geq 2$ and the CBF constraint set (or an inner approximation of it), $C(x)$, can be expressed as a hyperrectangle with lower and upper bounds $a(x), b(x) \in \mathbb{R}^n$, respectively, we maintain independent Beta distributions, $f^i(\cdot; \alpha^i_{\theta}(x), \beta^i_{\theta}(x))$, over each dimension $i$ of the unit box $[0, 1]^n$, and samples from these distributions are shifted and rescaled to lie within the bounds given by $a(x), b(x)$. Specifically, our CBF-constrained Beta policies, denoted $\pi_{\theta}$, sample $u \sim \pi_{\theta}(\cdot | x)$ from $C(x)$ by first sampling $\hat{u}^i \sim f^i(\cdot; \alpha^i_{\theta}(x), \beta^i_{\theta}(x))$, then performing the simple transformation $u = a(x) + \text{diag}( \hat{u}^1, \ldots, \hat{u}^n ) ( b(x) - a(x) )$, where $\text{diag}( \hat{u}^1, \ldots, \hat{u}^n )$ denotes the diagonal matrix with elements $\hat{u}^1, \ldots, \hat{u}^n$ along the diagonal.

\subsection{Implementation} \label{subsec:implementation}

We now describe the implementation details of our Beta policies. For a given state $x$, the parameter vectors $\alpha(x), \beta(x)$ are outputted by a two-layer, fully connected neural network. 
Control inputs at state $x$ were obtained by first creating an independent PyTorch \citep{paszke2019pytorch} Beta distribution object with parameters $\alpha^i_{\theta}(x), \beta^i_{\theta}(x)$, for each dimension $i \in \{1, \ldots, n\}$ of the action space, then sampling $u = [u^1 \ldots u^n ]^T$ from these distributions, and finally scaling and translating to lie within the current CBF set $C(x)$. Similarly, the Gaussian policies we used for comparison used distribution parameters outputted by a two-layer, fully connected neural network. Control inputs were then selected from the corresponding distribution by sampling, then following the standard practice \citep{chou2017improving} of clipping to a fixed set of permissible controls.
The PPO implementation used in the experiments was adapted with minor modifications from Stable Baselines 3 \citep{raffin2021stable}.

\subsection{Case study 1 : Quadcopter Navigation} \label{sec:case_study:quadcopter}


\textbf{Experiment Setup.}
For this experiment, we consider the problem of learning to safely navigate a quadcopter around an obstacle to a goal location. 
%
%
In this section, we present an overview of the dynamical model that we use for this quadcopter, which was previously considered in \cite{xu2018safe}, and describe our derivation of a hyperrectangular inner approximation of the safe control set, satisfying the CBF condition, that is amenable to sampling using our Beta policies. We finally briefly describe the reward function. \siva{See the supplementary material for a} detailed exposition of the environment and sampling procedure. 

We denote quadcopter and obstacle position by $r=(r_x,r_y,r_z)$ and $r_{obs}=(r_{o_x},r_{o_y},r_{o_z})$, respectively, and the quadcopter's relative position with respect to the obstacle as $ \Delta r=r-r_{obs}$. 
The quadrotor dynamics are then given by 
{\footnotesize\begin{align*}\label{eq:quad_dynamics}
\dot x=Ax+Bu, 
  x=  \begin{bmatrix}
        r\\
        \dot{r}
    \end{bmatrix}, 
  A  =
    \begin{bmatrix}
        \mathbf{0}_{3\times 3} & I_3\\
         \mathbf{0}_{3\times 3} & \mathbf{0}_{3\times 3}
    \end{bmatrix}, B=
    \begin{bmatrix}
        \mathbf{0}_{3\times 3}\\
        \mathbf{1}_{3\times 3}
    \end{bmatrix},
\end{align*}}
%
\noindent with input $u$ consisting the desired accelerations in each $x,y$, and  $z$ dimensions.
\siva{For the obstacle avoidance problem, we characterize the safe set as $\mathcal{S}=\{r: h(r)\geq 0\}$, where} 
\begin{equation}\label{eq:cbf_quad}
    h(r) = \left({\Delta r_{x}}/{a}\right)^4 + \left({\Delta r_{y}}/{b}\right)^4 + \left({\Delta r_{z}}/{c}\right)^4 -r_s,
\end{equation}
$a, b, c > 0$ parameterize the obstacle's shape, which is assumed to be elliptical, and $r_s$ represents the desired safety margin.
%
%
\siva{Given the quadcopter dynamics and \eqref{eq:cbf_quad},} 
the first time derivative $\dot{h}(r)$ does not explicitly contain \siva{the control input} $u$. We therefore \siva{use the standard ECBF formulation \cite{ames2019control} to} develop our safety condition using $\Ddot{h}(r)$, which explicitly contains $u$.  \siva{This ECBF condition is expressed as} 
$\Ddot{h} + K\cdot[h  \quad \dot{h}]^T \geq 0, $
where $K=[K_1 \quad K_2]^T, K_1 = 6, K_2 = 8$, are application-specific design parameters, \siva{and} can be rewritten as $A_r u \leq b_r,$ where $A_r$ is a matrix and $b_r$ is a vector, both depending on $r$. \siva{Then, the state-dependent} safe control set is given by $C(r) = \{ u \ : \ A_r u \leq b_r \}$ (see the supplementary for details). The dynamics (and consequently the ECBF conditions) are discretized with time step $dt=0.1$.
%
    


We consider navigation in the $x, y$ dimension as in \cite{xu2018safe}, resulting in a two-dimensional action space. We take the actuator constraint to be defined by the hyperrectangle $H:=\{u_{min}, u_{max}\} $, where $u_{min} := (u_{min}^x,u_{min}^y) \in \mathbb{R}^2$ and $u_{max} := (u_{max}^x,u_{max}^y) \in \mathbb{R}^2$ are the minimum and maximum input values.
In order to sample from the safe control set $C(r)$ at a given $r$ with our Beta policies, we need a hyperrectangular inner approximation. We obtain this inner approximation by formulating and solving a convex optimization problem yielding the highest volume hyperrectangle, $H_c(r)$, contained within $C(r)$.


Finally, we designed a reward providing an $\ell_2$ penalty based on agent distance from the goal, as well as a sizeable bonus for reaching the goal and a significant penalty for approaching the edge of the map. See the supplementary for details.

\textbf{Results.}
The experiments we conducted illustrate that safety-filter based approaches like those considered in \cite{cheng2019end} fail on simple cases of our safety-constrained quadcopter problem (see Figure \ref{fig:quadcopter_experiments_projection}), while our CBF-constrained Beta policy succeeds (Figure \ref{fig:quadcopter_experiments_beta}). For illustration purposes, Figures \ref{fig:quadcopter_experiments_beta} and \ref{fig:quadcopter_experiments_projection} present trajectories generated over the course of training. The corresponding learning curves are included in the supplementary material.
As illustrated in Figure \ref{fig:quadcopter_experiments_projection}, the safety-filter approach is effective at ensuring safety and also learns to successfully reach the goal when the obstacle is nonexistent or distant. However, it ultimately fails to reach the goal when the obstacle lies directly between the start and goal positions. We hypothesize that this is due to the fact that the projection-based approach attempts to learn an optimal policy for the unconstrained navigation problem, while projection causes it to deviate from its learned policy to maintain safety. Furthermore, the resulting safety-filtered policy cannot recover from these projections without an additional control layer (such as a derivative or PID controller) due to the repeated perturbation from the projection procedure.
%
%
Our method, on the other hand, learns to successfully solve the problem as shown in Figure \ref{fig:quadcopter_experiments_beta} while maintaining safety throughout training, since we directly learn policies for the CBF-constrained problem.
%


\subsection{Case study 2: Inverted pendulum}\label{sec:case_study:inverted_pendulum}

%
%

\begin{figure}[htb]
    \centering
    \includegraphics[width=0.94\linewidth]{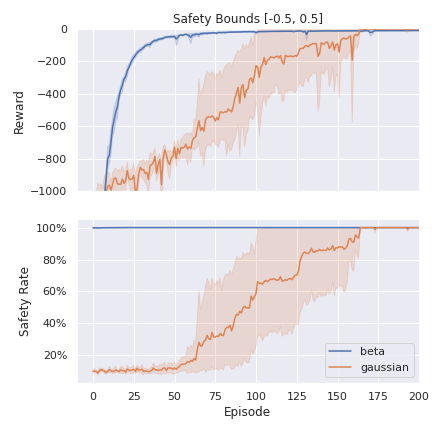}
    \vspace{-1em}
    \caption{CBF-constrained Beta vs. unconstrained Gaussian on inverted pendulum environment with safe set $\mathcal{S}_{0.5} = \{ \theta \ | \ -0.5 \leq \theta \leq 0.5 \}$. ``Safety Rate'' denotes percentage of time spent in safe set. Curves present mean and 95\% confidence intervals over 5 replications.}
    \label{fig:pendulum_experiments}
\end{figure}

\textbf{Experiment Setup.}
For the second set of experiments, we considered a safety-constrained inverted pendulum environment building on the baseline Gym implementation \citep{brockman2016openai}. The goal in this environment is to swing an inverted pendulum upright while maintaining it within a fixed safe set. We tested PPO with the two different policies on this environment for two different safe sets: $\mathcal{S}_{0.5} = \{ \theta \ | \ -0.5 \leq \theta \leq 0.5 \}$ and $\mathcal{S}_{1.0} = \{ \theta \ | \ -1.0 \leq \theta \leq 1.0 \}$. Due to space limitations, we include the experiments with $\mc{S}_{0.5}$ in Figure \ref{fig:pendulum_experiments} and the experiments with $\mc{S}_{1.0}$ with the supplementary material. As a baseline, we compare the proposed method to PPO with unconstrained Gaussian policies. This comparison highlights the effectiveness of the proposed method in guaranteeing safety as well as accelerating learning.

\textbf{Results.}
%
%
%
Our experiments are summarized in Figure \ref{fig:pendulum_experiments}. There are two main points to be drawn from these results.
First, the top panel shows that incorporating prior knowledge about properties such as safety can encourage learning and accelerate convergence
%
%
by forcing the Beta policy agent to concentrate on higher-value subsets of the state space. The Gaussian agent, on the other hand, is unable to benefit from this prior knowledge and convergence suffers as it spends a greater portion of its time exploring lower-value regions of the state space.
Second, the bottom panel illustrates that Beta policies are highly effective at maintaining safety throughout training, while Gaussian policies without safety constraints naturally fail to remain inside the safe set. This is expected, but illustrates the need to use constraint-aware policies such as Beta policies when prior knowledge is available. 

\section{CONCLUSION}

We have developed a sampling-based approach to learning policies ensuring hard constraint satisfaction in RL. Unlike existing, projection-based methods that ensure safety but lack convergence guarantees, our scheme provably does both. In addition to our theoretical contributions, we have also presented a practical solution method that leverages CBF-constrained Beta policies to ensure safety, and experimentally demonstrated its effectiveness on safe quadcopter navigation and inverted pendulum environments. Interesting directions for future work including extensions to the case where the constraint set must be estimated and application of our CBF-constrained Beta policies to real-world robotics problems.

%% file: AISTATS2024/sections/Acknowledgements.tex
\section*{Acknowledgements}

The authors would like to thank the anonymous reviewers for their helpful comments and Mostafa Mohamed Fa Abdelnaby of Purdue University for pointing out Remark \ref{rmk:2}.
The work of W. A. Suttle was supported by a Distinguished Postdoctoral Fellowship with the U.S. Army Research Laboratory.
V. K. Sharma was partially funded by a grant from the Purdue Engineering Initiative on Autonomous and Connected Systems.
The work of J. Liu was supported in part by the National Science Foundation (NSF) under grant 2230101, by the Air Force Office of Scientific Research (AFOSR) under award number FA9550-23-1-0175, and by U.S. Air Force Task Order FA8650-23-F-2603.
%
%
The work of K. C. Kosaraju and V. Gupta was partially supported by Army Research Office grants W911NF2310111, W911NF2310266, and W911NF-23-1-0316,  AFOSR grant F.10052139.02.005, Office of Naval Research grants F.10052139.02.009 and F.10052139.02.012, and NSF grant 2300355.

%% file: AISTATS2024/supplement.tex

%
%





%

%


\aistatstitle{Supplementary Material}

\section{Proofs}

\noindent \textbf{Proof of Proposition \ref{prop:irreducibility}.}
%
    We construct a sequence of $\varepsilon$-balls, each reachable from the previous element of the sequence, that leads from $x_0$ to $\mathcal{B}$, then show that the head of the Markov chain lies inside this sequence with positive probability. 
    Fix $\varepsilon > 0$ and let $\{ y_0, y_1, \ldots, y_N \} \subset \mathcal{S}$ be such that $B_{\varepsilon}(y_{k+1}) \subset R(B_{\varepsilon}(y_k))$, for $k = 1, \ldots, N-1$, and $B_{\varepsilon}(y_N) \cap \mathcal{B} \neq \emptyset$ (see Figure 1 for an illustration). For a given $\theta$, let $\{ x_n \}$ be the Markov chain induced on $\mathcal{S}$ by $\pi^C_{\theta}$ such that $x_0 = y_0$. We show that the trajectory $( x_0, x_1, \ldots, x_N )$ is contained within the set $\{ y_0 \} \times B_{\varepsilon}(y_1) \times \ldots \times B_{\varepsilon}(y_N)$ with strictly positive probability, which will imply that $\{ x_n \}$ enters $\mathcal{B}$ with strictly positive probability.
    For each $k = 1, \ldots, N$, consider the probability measure $\nu_k$ defined as
    $$\nu_k(S) = P(x \in S \ | \ x_{k-1}) = \int_{\mathcal{T}^{-1}_{x_{k-1}}(S)} \pi^C_{\theta}(a | x_{k-1}) \ da,$$
    for any $\mu$-measurable subset $S$ of $\mathcal{S}$. Note that $\nu_k$ is absolutely continuous with respect to $\mu$, written $\nu_k \ll \mu$, since $\mu(S) > 0$ if and only if $\mathcal{T}^{-1}_{x_{k-1}}(S) > 0$, by Assumption \ref{assum:volume_preservation}. The Lebesgue-Radon-Nikodym Theorem implies that there exists a $\mu$-integrable function $f_k : \mathcal{S} \rightarrow \mathbb{R}$, called the Radon-Nikodym derivative of $\nu_k$, such that $\nu_k(S) = \int_S f_k(x) dx$ (see \cite{folland1999real} for details). To make the link between $f_k$ and $\nu_k$ perfectly clear, let us write
    \begin{align*}
    f_k(x) &= \int_{\mathcal{T}^{-1}_{x_{k-1}}(x)} \pi^C_{\theta}(a | x_{k-1}) \ da, \\
    \nu_k(S) &= \int_S \int_{\mathcal{T}^{-1}_{x_{k-1}}(x)} \pi^C_{\theta}(a | x_{k-1}) \ da \ dx.
    \end{align*}
    By Assumptions \ref{assum:volume_preservation} and \ref{assum:positive_probability}, we also have $\mu \ll \nu_k$. Since both $\nu_k \ll \mu$ and $\mu \ll \nu_k$, the two measures are said to be {\em equivalent}, meaning that they agree on which sets have measure zero. Since $\mu$ and $\nu_k$ are equivalent, a standard result from real analysis allows us to take the Radon-Nikodym derivative $f_k$ to be strictly positive $\mu$-almost everywhere.
    As a first consequence, notice that
    \small
    \begin{align}
        P & \Big( (x_0, x_1, x_2) \in \{y_0\} \times B_{\varepsilon}(y_1) \times B_{\varepsilon}(y_2) \ | \ x_0 = y_0 \Big) \nonumber \\
        &= P \Big( (x_1, x_2) \in B_{\varepsilon}(y_1) \times B_{\varepsilon}(y_2) \ | \ x_0 = y_0 \Big) \cdot P(x_0 = y_0) \nonumber \\
        &= P \Big( (x_1, x_2) \in B_{\varepsilon}(y_1) \times B_{\varepsilon}(y_2) \ | \ x_0 = y_0 \Big) \nonumber \\
        &= \int_{B_{\varepsilon}(y_1)} \int_{T^{-1}_{x_1} (B_{\varepsilon}(y_2))} \pi^C_{\theta} (a_1 | x_1) f_1(x_1) \ da_1 \ dx_1 \label{eqn:positive1} \\
        &= \int_{B_{\varepsilon}(y_1)} \int_{T^{-1}_{x_1} (B_{\varepsilon}(y_2)} \pi^C_{\theta} (a_1 | x_1) \left[ \int_{T^{-1}_{x_0}(x_1)} \pi^C_{\theta} (a_0 | x_0) \ da_0 \right] \ da_1 \ dx_1. \nonumber
    \end{align}
    \normalsize
    Given Assumption \ref{assum:positive_probability}, equation \eqref{eqn:positive1} is strictly positive, since $f_1$ is strictly positive almost everywhere and the integrals are taken over sets of positive volume.
    
    Building on equation \eqref{eqn:positive1}, we have
    \small
    \begin{align}
        P \Big( (x_1, & \ldots, x_{N-1}, x_N) \in \\
        & \hspace{3mm} B_{\varepsilon}(y_1) \times \ldots \times B_{\varepsilon}(y_{N-1}) \times (B_{\varepsilon}(y_N) \cap \mathcal{B}) \ | \ x_0 = y_0 \Big) \label{eqn:positive2} \\
        = &\int_{B_{\varepsilon}(y_1)} \int_{T^{-1}_{x_1}(B_{\varepsilon}(y_2))} \pi^C_{\theta}(a_1 | x_1) \cdot \int_{B_{\varepsilon}(y_2)} \int_{T^{-1}_{x_2}(B_{\varepsilon}(y_3))} \pi^C_{\theta}(a_2 | x_2) \cdot \ldots  \nonumber \\
        & \hspace{3mm} \ldots \cdot \int_{B_{\varepsilon}(y_{N-1})} \int_{T^{-1}_{x_{N-1}}(B_{\varepsilon}(y_N) \cap \mathcal{B})} \pi^C_{\theta}(a_{N-1} | x_{N-1}) \ f_{N-1}(x_{N-1}) \ da_{N-1} \ dx_{N-1} \cdot \ldots \nonumber \\
        & \hspace{12mm} \ldots \cdot f_2(x_2) \ da_2 \ dx_2 \cdot f_1(x_1) \ da_1 \ dx_1. \nonumber
    \end{align}
    \normalsize
    Note in the innermost integral that $\mu(B_{\varepsilon}(y_N) \cap \mathcal{B}) > 0$, since both sets are open and their intersection is non-empty by hypothesis. Finally, given Assumption \ref{assum:positive_probability}, we have that \eqref{eqn:positive2} is strictly positive, since all integrals are taken over sets of positive volume and $f_i$ is strictly positive almost everywhere, for each $i \in \{1, \ldots N-1\}$.
%
\hfill $\square$

\noindent \textbf{Proof of Lemma \ref{lemma:piC_assumptions_hold}.}
%
%
Recalling the definition of $\pi^C_{\theta}(u | x)$ in \eqref{eqn:piC_definition},
\begin{align}
    \nabla \log \pi^C_{\theta}(u | x) &= \nabla \log \pi_{\theta}(u | x) - \nabla \log \int_{C(x)} \pi_{\theta}(w | x) dw \nonumber \\
    &= \nabla \log \pi_{\theta}(u | x) - \frac{ \int_{C(x)} \nabla \pi_{\theta}(w | x) dw }{ \int_{C(x)} \pi_{\theta} (w | x) dw }. \label{eqn:piC_grad}
\end{align}

To see that, for all $x \in \mathcal{X}, u \in \mathcal{U}$, $\nabla \log \pi^C_{\theta}(u | x)$ exists, for all $\theta \in \Theta$, we simply need to verify that $\large(\int_{C(x)} \pi_{\theta}(w | x) dw \large)^{-1}$ is always finite. But this follows immediately from Assumption \ref{assum:positive_probability} and the fact that $\mu(C(x)) \geq m > 0$ by Assumption \ref{assum:muC_bounded}.

We next prove part 1) of the Lemma. The claim holds for the first term in \eqref{eqn:piC_grad} by part 2a) of Assumption \ref{assum:r_and_pi}, so we just need to show that it holds for the second term. To do this, we prove that, for a given $x \in \mathcal{X}$, this term is Lipschitz in $\theta$, then argue that the largest minimal Lipschitz constant over all $x \in \mathcal{X}$ is finite.
We know by part 2b) of Assumption \ref{assum:r_and_pi} that, for all $x \in \mathcal{X}, u \in \mathcal{U}$, $\norm{\nabla \pi_{\theta}(u | x)} \leq \nabla \log \pi_{\theta}(u|x) \leq B_{\Theta}$, for all $\theta \in \Theta$. This means that, for all $x \in \mathcal{X}$,
\begin{align*}
    \Big| \int_{C(x)} &\pi_{\theta}(w | x) dw - \int_{C(x)} \pi_{\theta'} (w | x) dw \Big| \\
    &= \Big| \int_{C(x)} \left( \pi_{\theta}(w | x) - \pi_{\theta'} (w | x)  \right) dw \Big| \\
    &\leq \int_{C(x)} | \pi_{\theta}(w | x) - \pi_{\theta'} (w | x) | dw \\
    &\leq \int_{C(x)} B_{\Theta} \norm{ \theta - \theta' } dw = B_{\Theta} \mu(C(x)) \norm{ \theta - \theta'} \\
    &\leq B_{\Theta} M \norm{\theta - \theta'},
\end{align*}
for all $\theta, \theta' \in \Theta$. So $\int_{C(x)} \pi_{\theta}(w | x) dw$ is Lipschitz in $\theta$, for each $x \in \mathcal{X}$, and the largest Lipschitz constant over $\mathcal{X}$ is finite. In addition, $\int_{C(x)} \pi_{\theta}(w | x) dw$ is clearly uniformly bounded.

Notice that $\inf_{x \in \mathcal{X}} \mu(C(x)) \geq m > 0$, by Assumption \ref{assum:muC_bounded}. Thus, by Assumption \ref{assum:positive_probability}, $\inf_{x \in \mathcal{X}} \int_{C(x)} \pi_{\theta}(w | x) dw > 0$, for all $\theta \in \Theta$. Since $\Theta$ is compact, this means $\inf_{\theta \in \Theta} \inf_{x \in \mathcal{X}} \int_{C(x)} \pi_{\theta}(w | x) dw > 0$. This implies that $\large(\int_{C(x)} \pi_{\theta}(w | x) dw \large)^{-1}$ is uniformly bounded. Since $\int_{C(x)} \pi_{\theta}(w | x) dw$ is Lipschitz, $\large(\int_{C(x)} \pi_{\theta}(w | x) dw \large)^{-1}$ is therefore Lipschitz and bounded in $\theta \in \Theta$, for all $x \in \mathcal{X}$. We also know that, for each $x \in \mathcal{X}$, $\int_{C(x)} \nabla \pi_{\theta}(w | x) dw$ is Lipschitz and bounded in $\theta \in \Theta$, by Assumption \ref{assum:r_and_pi}, part 2a). Fix $x \in \mathcal{X}$. Since the product of Lipschitz, bounded functions is Lipschitz and bounded, the function $\int_{C(x)} \nabla \pi_{\theta}(w | x) dw / \int_{C(x)} \pi_{\theta}(w | x) dw$ is Lipschitz and bounded in $\theta$. Since this function is uniformly bounded over $x \in \mathcal{X}, \theta \in \Theta$, there therefore exists $L > 0$ such that, for all $x \in \mathcal{X}$,
\begin{equation*}
    \norm{ \frac{ \int_{C(x)} \nabla \pi_{\theta}(w | x) dw }{ \int_{C(x)} \pi_{\theta}(w | x) dw } -\frac{ \int_{C(x)} \nabla \pi_{\theta'}(w | x) dw }{ \int_{C(x)} \pi_{\theta'}(w | x) dw } } \leq L \norm{ \theta - \theta' },
\end{equation*}
for all $\theta, \theta' \in \Theta$. Combined with part 2a) of Assumption \ref{assum:r_and_pi}, this implies that, for all $x \in \mathcal{X}, u \in \mathcal{U}$, $\norm{ \nabla \log \pi^C_{\theta}(u | x) - \nabla \log \pi^C_{\theta'}(u | x)} \leq (L_{\Theta} + L) \norm{\theta - \theta'}$, for all $\theta, \theta' \in \Theta$. This completes the proof of part 1).

Part 2) follows from the fact that $\int_{C(x)} \nabla \pi_{\theta}(w | x) dw / \int_{C(x)} \pi_{\theta}(w | x) dw$ is uniformly bounded and that, for all $x \in \mathcal{X}, u \in \mathcal{U}$, $\norm{ \nabla \log \pi_{\theta}(u | x) } \leq B_{\Theta}$, for all $\theta \in \Theta$, by part 2) of Assumption \ref{assum:r_and_pi}.
%
%

\section{Background: Barrier Functions}\label{appendix:subsection:cbf}
In this section, we provide an overview of barrier functions for convenience. The theory of barrier functions revolve around controlled set invariance for dynamical systems. Safety can be represented through a set, say $\mathcal{S}$, defined as a level set of this barrier function, say $h$. Then, we write the condition on this barrier function to guarantee the forward invariance of this safety set under the given dynamics.

\subsection{Control Barrier Functions (CBF)}
Consider the following nonlinear system
\begin{align}\label{appendix:control_barrier_funs_system}
    \dot{r} =f(r,u),
\end{align}
where $r \in D \subset \mathbb{R}^n$ and $u \in U \subset \mathbb{R}^m$ denote the state and control input, and $f$ is a locally  Lipschitz function that models the state transition. The following definition and the theorem follows the development in ~\cite{ames2019control, agrawal2017discrete}. 

\begin{theorem}\label{appndix:cbf:theorem:cbf} 
    Consider a function $h:\mathbb{R}^{n} \rightarrow \mathbb{R}$ that is continuously differentiable. Define a closed set $\mathcal{S}$ as the super-level set of this function as follows:
    \begin{align}\label{set:forward_inv}
        \mathcal{S} \triangleq \left\{r\in \mathbb{R}^{n} \;| \;h(r)\geq 0\right\}.
    \end{align}
    The function $h$ is a control barrier function, for \eqref{appendix:control_barrier_funs_system} and with state $s$, if there exists an extended $\kappa_\infty$ function  $\alpha$ such that for all $r\in \mathcal{S}$, $t\in \mathbb{R}_+$,
    \begin{equation}\label{ctrl:deterministic_pol}
           \dot{h} \geq  -\alpha(h).
    \end{equation}
    Further, if we define the safe control set as
    \begin{equation}
        \mathcal{C}(r) \triangleq \left\{u\in \mathbb{R}^m | \dot{h}(r,u) \geq  -\alpha(h(r))\right\}.
    \end{equation}
    then any input $u\in \mathcal{C}(r)$  will render the set $\mathcal{S}$ forward invariant. 
\end{theorem}

When designing safe controller with control values $u$ sampled from this safe control set, we need the time-derivative of $h$ i.e. $\dot{h}$ to explicitly contain $u$. However, the above forward variance condition is restricted to barrier function with relative degree $d_r=1$.
At this point, we also note that for our quadcopter experiment, our CBF has a relative degree $d_r=2$, since only the second time-derivative $\Ddot{h}$ explicitly contains the control input $u$. 
 Therefore, for barrier functions with relative degree more than $1$, which are often referred to as Exponential Control Barrier Functions (ECBF), we need a seperate discussion on forward invariance conditions.

\subsection{Exponential Control Barrier Functions}
We now discuss exponential CBFs for control affine nonlinear dynamical system. Consider the following control affine nonlinear dynamical system:
\begin{equation}\label{eq:affine-dyn}
    \dot{r} = f(r) + g(r) u,
\end{equation}
with $f$ and $g$ locally lipshitz, $r \in D \subset \mathbb{R}^n$ and $u \in U \subset \mathbb{R}^m$. We suppose that the Lipschitz constant for $f$ and $g$ are $L_f$ and $L_g$ respectively, and the vector containing the first $d_r-1$ time derivatives of $h(r)$ including $h(r)$ is given as:
    $\eta_b(r)=
    \begin{bmatrix}
       h(r)\\
       \dot{h}(r)\\
       \Ddot{h}(r)\\
       \vdots \\
       h^{d_r-1}(r)
    \end{bmatrix}
    $. Further suppose that the matrices $F, G,$ and $C$ are defined as follows:
    $F=
    \begin{bmatrix}
       0 & 1 & 0 & \cdots & 0\\
       0 & 0 & 1 & \cdots & 0\\
       \vdots & \vdots & \vdots & \ddots & \vdots  \\
       0 & 0 & 0 & \cdots & 1 \\
        0 & 0 & 0 & \cdots & 0
    \end{bmatrix}
    $,
    $G=
    \begin{bmatrix}
       0\\
       0\\
       0\\
       \vdots \\
       1
    \end{bmatrix}
    $, and,
    $C=
    \begin{bmatrix}
       1 & 0 & 0 & \cdots & 0
       \end{bmatrix}$.

\begin{theorem}
    Consider a function $h:\mathbb{R}^{n} \rightarrow \mathbb{R}$ that is continuously differentiable. Define a closed set $\mathcal{S}$ as the super-level set of this function as follows:
    \begin{align}\label{set:forward_inv}
        \mathcal{S} \triangleq \left\{r\in \mathbb{R}^{n} \;| \;h(r)\geq 0\right\}.
    \end{align}
    Then the function $h$ is an ECBF, with relative degree $d_r$, for system in \eqref{eq:affine-dyn}, if there exist a row vector $K_\alpha \in \mathbb{R}^r$ such that 
    \begin{equation}
        \sup_{u \in U} [L_f^{d_r}h(r) + L_g L_f^{d_r-1}h(r)u] \geq - K_\alpha \eta_b(r)
    \end{equation}
    $\forall r \in Int(\mathcal{S})$,
    implies $h(r(t))\geq Ce^{(F-GK_{\alpha})t} \eta_b(r) r(t_0) \geq 0$, whenever $h(r(t_0))\geq 0$.
    Further, if we define the safe control set $C(r)$ as
    \begin{equation}
        \mathcal{C}(r) \triangleq \{u \in U | [L_f^{d_r}h(r) + L_g L_f^{d_r-1}h(r)u] \geq - K_\alpha \eta_b(r)\},
    \end{equation}
    then any input $u\in \mathcal{C}(r)$ will render the set $\mathcal{S}$ forward invariant.
\end{theorem}

\section{Experiments: Additional Details}

In this section, we provide additional details regarding the experiments presented in \S\ref{sec:experiments}.

\subsection{Inverted Pendulum Experiments} 

The safety-constrained inverted pendulum environment that we considered in \S\ref{sec:case_study:inverted_pendulum} was obtained by modifying the standard implementation from \cite{brockman2016openai} to include CBF-based safety constraints. In this section we describe the dynamical model and CBF used to obtain these constraints, then present implementation details and an additional experiment. 

\subsubsection{Dynamical Model}

Consider the model of a simple inverted pendulum
\begin{align} \label{case_study:inverted_pendulum:dynamics}
    \begin{bmatrix}
    \theta_{k+1}\\
    \dot \theta_{k+1}
    \end{bmatrix}
&=
    \begin{bmatrix}
    \theta_{k}+\delta t\dot \theta_k+\delta t^2 \left(\dfrac{3g}{2l}\sin{\theta_k}+\dfrac{3}{ml^2}u_k\right)\\
    \dot \theta_{k}+\delta t \left(\dfrac{3g}{2l}\sin{\theta_k}+\dfrac{3}{ml^2}u_k\right)
    \end{bmatrix},
\end{align}
where $\theta_k,~\dot \theta_k$ denote the states (angle and angular velocity), $u_k$ denote the input (torque),  $m$ and $l$ denotes the mass and the length of the pendulum, respectively,  $g$ denotes the acceleration due to gravity and $\delta t>0$ denotes the discretization time.  Denote the safe operating region by
\begin{align}\label{case_study:inverted_pendulum:safeset}
    \mathcal{S}&=\left\{\theta\in \mathbb{R}|h(\theta):=\begin{bmatrix}
    \theta+1\\ 1-\theta
    \end{bmatrix}\geq 0\right\}.
\end{align}


\subsubsection{Control Barrier Function}

The following corollary is a direct consequence of Theorem \ref{appndix:cbf:theorem:cbf}, presented in \S\ref{appendix:subsection:cbf}.
\begin{corollary}\label{case_study:inverted_pendulum:corollary:safety}
Let
    \begin{align}\label{case_study:inverted_pendulum:safe_action}
   & U(\theta_k, \dot \theta_k) = \left\{u_k \in \bR |
    \left(\delta t\dot \theta_k+ c(\theta_k, u_k)\right)\begin{bmatrix}
    1\\ -1
    \end{bmatrix} + \eta \begin{bmatrix}
    \theta_k+1\\ 1-\theta_k
    \end{bmatrix} \geq 0
    \right\}
\end{align}
where $c(\theta_k, u_k):=\delta t^2\left(\dfrac{3g}{2l}\sin{\theta_k}+\dfrac{3}{ml^2}u_k\right)$, and $0<\eta<1$. Consider system \eqref{case_study:inverted_pendulum:dynamics} with $u_k\in U(\theta_k, \dot \theta_k)$. Let $(\theta_{0}, \dot{\theta}_0) \in \mathcal{S}\times \mathbb{R}$ and assume $U(\theta_0, \dot{\theta}_0)$ is non-empty. The set $\mathcal{S}$ is forward invariant.
\end{corollary}
The safe set \eqref{case_study:inverted_pendulum:safe_action} is used to provide the state-dependent constraints to the Beta policies learned in our experiments.

\subsubsection{Implementation Details}

We next describe the implementation details of our experiments. As mentioned above, the environment was adapted from the implementation of \citep{brockman2016openai}, with modifications to compute the CBF safe set \eqref{case_study:inverted_pendulum:safe_action}. The reward function and other details are as in \cite{brockman2016openai}. The Beta and Gaussian policies used the corresponding distributions from the PyTorch library \citep{paszke2019pytorch}. As described in \S\ref{subsec:implementation}, for a given state $x$, the parameters $\alpha(x), \beta(x)$ of the Beta distribution were outputted by a two-layer, fully connected neural network. Control inputs were obtained by sampling from this distribution, then translating and rescaling to lie within the current CBF set $C(x) = [a(x), b(x)]$. The Gaussian policy parameters were outputted by a two-layer, fully connected neural network. Control inputs were subsequently selected from the corresponding distribution by sampling, then, following standard practice \citep{chou2017improving}, were clipped to a set of permissible controls, which was chosen to be $[-15.0, 15.0]$. The hyperparameters used are presented in Figure \ref{fig:pendulum_hyperparameters}.



\begin{figure}
    \begin{subfigure}[t]{0.49\textwidth}
        \begin{center}
            \begin{tabular}{ |c|c| }
                 \hline
                 policy learning rate & 0.0003 \\ 
                 \hline
                 value learning rate & 0.0003 \\
                 \hline
                 entropy coefficient & 0.0 \\
                 \hline
                 clip range & 0.2 \\
                 \hline
                 weight decay & 0.0 \\
                 \hline
                 layer size & 64 \\
                 \hline
                 batch size & 64 \\
                 \hline
                 buffer size & 300 \\
                 \hline
                 number of epochs & 10 \\
                 \hline
                 rollout length & 300 \\
                 \hline
                 discount factor & 0.99 \\
                \hline
            \end{tabular}
        \end{center}
        \caption{Gaussian hyperparameters.}
    \end{subfigure}
    \hfill
    \begin{subfigure}[t]{0.49\textwidth}
        \begin{center}
            \begin{tabular}{ |c|c| }
                 \hline
                 policy learning rate & 0.01 \\ 
                 \hline
                 value learning rate & 0.01 \\
                 \hline
                 entropy coefficient & 0.0 \\
                 \hline
                 clip range & 0.2 \\
                 \hline
                 weight decay & 0.0 \\
                 \hline
                 layer size & 64 \\
                 \hline
                 batch size & 64 \\
                 \hline
                 buffer size & 300 \\
                 \hline
                 number of epochs & 10 \\
                 \hline
                 rollout length & 300 \\
                 \hline
                 discount factor & 0.99 \\
                \hline
            \end{tabular}
        \end{center}
        \caption{Beta hyperparameters.}
    \end{subfigure}
    \caption{PPO hyperparameters for the inverted pendulum experiments.}
    \label{fig:pendulum_hyperparameters}
\end{figure}

\subsubsection{Additional Results}

%
\begin{figure}
    \centering
    \includegraphics[width=0.5\textwidth]{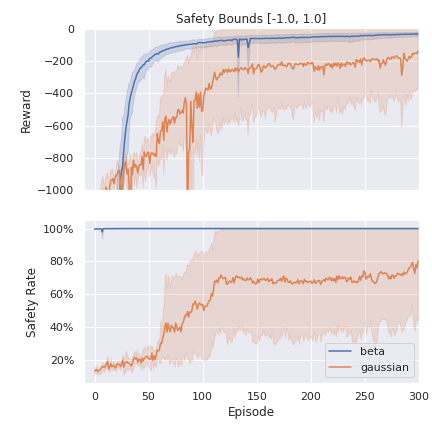}
    \caption{Comparison of safety-constrained Beta policy and unconstrained Gaussian policy on the inverted pendulum environment with constraint set $\mathcal{S}_{1.0} = \{ \theta \ | \ -1.0 \leq \theta \leq 1.0 \}$. The top figure presents learning curves, while the bottom figure presents the ``safety rate'', i.e., the percentage of time spent in $\mathcal{S}_{1.0}$ over the course of the episode. The curves represent means and 95\% confidence intervals over five independent replications.}
    \label{fig:theta_bds_1.0}
\end{figure}

Figure \ref{fig:theta_bds_1.0} presents an experiment providing additional support to the discussion presented in \S\ref{sec:case_study:inverted_pendulum}.

\subsection{Quadcopter Experiments}

In this section, we provide additional details regarding the environment and experiments presented in \S\ref{sec:case_study:quadcopter}.

\subsubsection{Dynamical Model}
We summarize the dynamical model of the quadcopter derived in \cite{xu2018safe}. We consider the body frame, say $\mathbb{F}_b$, and world frame, say $\mathbb{F}_w$, and discuss the transformation between these two frames using the rotation matrix $\mathbb{R}_{wb}$ defined as
\begin{equation}\label{eq:rotation}
    \mathbb{R}_{wb}:=
    \begin{bmatrix}
    \cos{\psi}\cos\theta-\sin{\phi}\sin{\psi}\sin{\theta} & -\cos{\phi}\sin{\psi} & \cos{\psi}\sin{\theta}+\cos{\theta}\sin{\phi}\sin{\psi}\\
    \cos{\theta}\sin{\psi}+\cos{\psi}\sin{\phi}\sin{\theta} & \cos{\phi}\cos{\psi} &  \sin{\psi}\sin{\theta}-\cos{\psi}\cos{\theta}\sin{\phi}
    \\
    -\cos{\phi}\sin{\theta} & \sin{\phi} & \cos{\phi}\cos{\theta}
    \end{bmatrix},
\end{equation}
where $\phi$, $\theta$, and  $\psi$ denote the Z-X-Y Euler angles corresponding to the roll, pitch, and yaw of the quadcopter. Suppose that the 3-dimensional position coordinates of the quadcopter along the x-,y-, and z-axis with respect to its body frame $\mathbb{F}_b$ of and the world frame of reference  $\mathbb{F}_w$ be given by $x_b:=(x_b,y_b,z_b)$ and $r:=(r_x,r_y,r_z)$ respectively, then $r=\mathcal{R}_{wb}x_b$. 

Then, the quadcopter dynamics is given by $\dot{x}=Ax+Bu$, where the control input $u$ comprises of the desired acceleration of the quadcopter. The dynamics of this controller under small angle assumptions on the Euler angles, that is $\sin \hat e\approx \hat e, \cos \hat e\approx 1, \hat e\in \{\phi, \theta, \psi\}$) evolves as  \cite{mellinger2012trajectory}: 

\vspace{-3mm}
{\small
\begin{align} \label{ang-to-r}
    u=
    \begin{bmatrix}
       \Ddot{r}_x^{des}\\
       \Ddot{r}_y^{des}\\
       \Ddot{r}_z^{des}
    \end{bmatrix}=
    \begin{bmatrix}  g(\theta^{des}\cos{\psi^{des}}+\phi^{des}\sin{\psi^{des}}), \\ 
   g(\theta^{des}\sin{\psi^{des}}-\phi^{des}\cos{\psi^{des}}) \\
    \frac{ \sum^4_{i=1}F_i^{des}}{m} -g \end{bmatrix},
\end{align}}
\vspace{-3mm}

\noindent where $m, g$ are respectively the mass of the quadcopter and gravitational constant, and $\Ddot{r}_i^{des}, i\in \{x,y,z\}$ is the desired acceleration component of the quadcopter in the x-,y-, and z-direction respectively, computed using the desired specifications on the Euler angles $\phi^{des}, \theta^{des}$, and $\psi^{des}$, and $F_i^{des},i\in \{1,2,3,4\}$ is the desired thrust on the $i$-th rotor of the quadcopter. 
Lastly, the dynamical parameters for the quadcopter are setup as given in \cite{hoshih2020provablyinwild}.


\subsubsection{Exponential Control Barrier Function}
Recall, that the objective of our controller to enable the quadcopter to learn how to reach a target position $r_{goal}$, while avoiding an obstacle with  position $r_{obs}$. For this obstacle avoidance, we now discuss our choice of CBF for the quadcopter experiment, as defined in \eqref{eq:cbf_quad}, and reason why this is an exponential control barrier function (ECBF). We first derive expressions for $\dot{h}$ and $\Ddot{h}$ using dynamical equations as follows:

\begin{equation} 
    \dot{h}(r) = 4(\left({\Delta r_{x}}/{a}\right)^3 \dot{r}_x + \left({\Delta r_{y}}/{b}\right)^3 \dot{r}_y + \left({\Delta r_{z}}/{c}\right)^3 \dot{r}_z),
\end{equation}
and
\begin{equation}
    \Ddot{h}(r) = 12(\left({\Delta r_{x}}/{a}\right)^2 \dot{r}_x + \left({\Delta r_{y}}/{b}\right)^2 \dot{r}_y + \left({\Delta r_{z}}/{c}\right)^2 \dot{r}_z) + 4(\left({\Delta r_{x}}/{a}\right)^3 \Ddot{r}_x + \left({\Delta r_{y}}/{b}\right)^3 \Ddot{r}_y + \left({\Delta r_{z}}/{c}\right)^3 \Ddot{r}_z).
\end{equation}

These equations can be re-written in vector form as follows:
\begin{equation}\label{eq:hdot}
    \dot{h}(r) = \begin{bmatrix}
        4({\Delta r_{x}^3}/{a}^4) & 4({\Delta r_{y}^3}/{b}^4) & 4({\Delta r_{z}^3}/{c}^4)
    \end{bmatrix}\dot{r}
\end{equation}

and
\begin{equation}
    \Ddot{h}(r) = \dot{r}^T \begin{bmatrix}
        12({\Delta r_{x}}^2/{a}^4) & 0 & 0\\
        0 & 12({\Delta r_{y}^2}/{b}^4) & 0\\
        0 & 0& 12({\Delta r_{z}^2}/{c}^4)
    \end{bmatrix}\dot{r}
    + \begin{bmatrix}
        4({\Delta r_{x}}^3/{a}^4) & 4({\Delta r_{y}^3}/{b}^4) & 4({\Delta r_{z}^3}/{c}^4)
    \end{bmatrix}\Ddot{r}.
\end{equation}

Since $u = \Ddot{r}$ from quadcopter dynamics, we can re-write the $\Ddot{h}(r)$ as follows:
\begin{equation}\label{eq:hddot}
    \Ddot{h}(r) = \dot{r}^T D_r \dot{r}
     -A_ru,
\end{equation}
where 
$D_r := \begin{bmatrix}
        12({\Delta r_{x}}^2/{a}^4) & 0 & 0\\
        0 & 12({\Delta r_{y}^2}/{b}^4) & 0\\
        0 & 0& 12({\Delta r_{z}^2}/{c}^4)
    \end{bmatrix}$, and
    $A_r:=-
    \begin{bmatrix}
        4({\Delta r_{x}}^3/{a}^4) & 4({\Delta r_{y}^3}/{b}^4) & 4({\Delta r_{z}^3}/{c}^4)
    \end{bmatrix}$.

Therefore, we note that $\Ddot{h}(r)$ or the $2^{nd}$ time-derivative of $h(r)$ which explicitly depends on the control input $u$ and therefore our choice of CBF $h(r)$ is an exponential CBF with a relative degree \cite{ames2019control} of $2$.

Correspondingly, we use the following forward invariance condition for the set $\mathcal{S} = \{r : h(r) \geq 0\}$ as given in \cite{xu2018safe}:
\begin{equation}\label{eq:ecbf}
    \Ddot{h} + K\cdot[h  \quad \dot{h}]^T \geq 0,
\end{equation}
with $K=[K_1 \quad K_2]^T$.

The above equation can be re-arranged as follows:
\begin{equation*}
    -\Ddot{h} \leq K_1h +K_2\dot{h},
\end{equation*}
and, using \eqref{eq:hdot} and \eqref{eq:hddot}, we can re-write this equation as
\begin{equation}\label{eq:ecbf-short}
    A_ru \leq b_r, 
\end{equation}
where
$b_r=\dot{r}^T D_r \dot{r} + K_1 h - K_2A_r\dot{r}$. Thus, we can write the safe control set as $\mathcal{C}(r) = \{u \in \mathbb{R}^3: A_r u \leq b_r \}$. For our quadcopter experiments, we consider navigation $x-y$ dimensions, therefore set the $z$-dimension position and velocity to be $0$. Thus the control input only comprises the desired acceleration for $x$ and $y$ axes and therefore our action space becomes two-dimensional, and we only consider the $x$ and $y$ components in the above CBF calculations.

\subsubsection{Maximal Inner Hyperrectangle Computation} 

We now describe the construction of the maximal inner hyperrectangle contained in the set $\mathcal{C}(r)$ under actuator constraints $H$. These are the sets that our Beta policies will sample from.
We use the following optimization problem, with decision variables $u=(u^x,u^y)$, to get the maximal inner hyper-rectangle inside the safe set:
\begin{equation}\label{opt:inner_approximation}
    \mathcal{P}_A :\begin{aligned}
        \underset{u}{\mathrm{max}}  
         \ \ &\mathcal{A}(u) \\
        \text{s.t.} \ \  &A_ru \leq b_r, \\ &u \in H,
    \end{aligned} 
\end{equation}
where $\mathcal{A}(u)$ is the area of a hyperrectangle inside $C(r)\cap H$ and the decision variables $u^x,u^y$ are points on the line $A_ru\leq b_r$. 
One of the corner points of this hyperrectangle is formed by $u^x,u^y$ and the rest of corner points lie on the boundary hyperrectangle formed by $H$.
Suppose that $(u^x_*,u^y_*)$ are solutions to $\mathcal{P}_A$, then the definition of Area $\mathcal{A}$ depends on how the line $A_ru\leq b_r$ intersects with $H$, and therefore, leads to the following four possibilities:
\begin{itemize}
    \item $\mathcal{A} = (u^x-u_{min}^x)*(u^y-u_{min}^x)$ and $H_c = \{u_{min},(u^x_*,u^y_*)\}$
    \item $\mathcal{A} = (u^x-u_{min}^x)*(u_{max}^y-u^y)$ and $H_c = \{(u_{min}^x,u^y_{*}),(u^x_{*},u_{max}^y)\}$
    \item $\mathcal{A} = (u_{max}^x - u^x)*(u_{max}^x -u^y)$ and $H_c = \{(u^x_{*},u^y_{*}),u_{max}\}$
    \item $\mathcal{A} = (u_{max}^x-u^x)*(u^y-u_{min}^y)$ and $H_c = \{(u^x_*,u_{min}^y),(u_{max}^x,u^y_*)\}$.
\end{itemize}

$\mathcal{P}_A$ is, in general, a non-convex program. However, through change-of-variables, we can transform this problem into a tractable problem through the following transformation.
%
%
We perform a change of variables, with new variables denoted by $(\bar{u}^x,\bar{u}^y)$ defined by
\begin{itemize}
    \item $\bar{u}^x = u^x-u_{min}^x$ and $\bar{u}^y = u^y-u_{min}^y$
    \item $\bar{u}^x = u^x-u_{min}^x$ and $\bar{u}^y = u_{max}^y-u^y$
    \item $\bar{u}^x = u_{max}^x-u^x$ and $\bar{u}^y = u_{max}^y-u^y$
    \item $\bar{u}^x = u_{max}^x-u^x$ and $\bar{u}^y = u^y-u_{min}^y$.
\end{itemize}
The corresponding objective function is given by $\mathcal{\bar{A}}= \bar{u}^x\bar{u}^y$.
%
%
So long as the entries corresponding to $A_r$ and $b_r$ from \eqref{opt:inner_approximation} in the transformed problem are nonnegative,
the resulting problem is a geometric program, which can be further transformed to a convex problem by standard methods and efficiently solved. In our quadcopter experiments, when $A_r, b_r \geq 0$, we solved the transformed geometric program using CVXPY \citep{diamond2016cvxpy}, and used non-linear solvers from SCIPY \citep{virtanen2020scipy} otherwise. We observed in our experiments that the transformation resulted in geometric programs in all but a handful of cases.
%

\subsubsection{Reward}
We now discuss reward shaping used in our quadcopter experiments.
%

Suppose that the $r_{min}:=[r_{min}^x \quad r_{min}^y]^T$ and $r_{max}:=[r_{max}^x \quad r_{max}^y]^T$ are environment boundaries, with $x, y$-axis boundary repectively defined by $[r_{min}^x,r_{max}^x]$ and  $[r_{min}^y,r_{max}^y]$, that we employ for guiding exploration for both the quadcopter experiments. Then the reward used in our environment is defined by
\begin{equation*}
R(r)=
    \begin{cases}
        50 & \text{ if } ||r-r_{goal}||_2 < \epsilon,\\
        -||r-r_{goal}||_2 & \text{ if } r_{max} > r > r_{min} \text{ and } ||r-r_{goal}||_2 \geq \epsilon,\\
        -||r-r_{goal}||_2 - 400 & \text{ if } r \geq r_{max} \text{ or } r \leq r_{min},
    \end{cases}
\end{equation*}
where $\epsilon = 0.25$ is the boundary around $r_{goal}$ for which we give a constant positive reward of $50$ to the agent, and the inequalities in the reward definition are element-wise. Moreover, when the agent is inside the boundary but outside the $\epsilon$-neighborhood of the goal, then the reward is negative of the distance between the agent and the goal. Lastly, we penalize the agent if it goes outside the boundary defined by $r_{min}$ and $r_{max}$ to encourage exploration in the region around the goal.

\subsubsection{Hyperparameters}

The hyperparameters used in the experiments are presented in Figure \ref{fig:quadcopter_hyperparameters}.

\begin{figure}
    \begin{subfigure}[t]{0.49\textwidth}
        \begin{center}
            \begin{tabular}{ |c|c| }
                 \hline
                 policy learning rate & 0.0004 \\ 
                 \hline
                 value learning rate & 0.0004 \\
                 \hline
                 entropy coefficient & 0.00000001 \\
                 \hline
                 clip range & 0.2 \\
                 \hline
                 weight decay & 0.0 \\
                 \hline
                 layer size & 256 \\
                 \hline
                 batch size & 256 \\
                 \hline
                 buffer size & 320 \\
                 \hline
                 number of epochs & 10 \\
                 \hline
                 rollout length & 320 \\
                 \hline
                 discount factor & 0.90 \\
                \hline
            \end{tabular}
        \end{center}
        \caption{Gaussian hyperparameters.}
    \end{subfigure}
    \hfill
    \begin{subfigure}[t]{0.49\textwidth}
        \begin{center}
            \begin{tabular}{ |c|c| }
                 \hline
                 policy learning rate & 0.0006 \\ 
                 \hline
                 value learning rate & 0.0006 \\
                 \hline
                 entropy coefficient & 0.0 \\
                 \hline
                 clip range & 0.2 \\
                 \hline
                 weight decay & 0.0 \\
                 \hline
                 layer size & 256 \\
                 \hline
                 batch size & 256 \\
                 \hline
                 buffer size & 180 \\
                 \hline
                 number of epochs & 10 \\
                 \hline
                 rollout length & 180 \\
                 \hline
                 discount factor & 0.90 \\
                \hline
            \end{tabular}
        \end{center}
        \caption{Beta hyperparameters.}
    \end{subfigure}
    \caption{PPO hyperparameters for the quadcopter experiments.}
    \label{fig:quadcopter_hyperparameters}
\end{figure}

\subsubsection{Learning Curves}

\begin{figure*}[htb]
     \centering
     \begin{subfigure}[t]{0.49\textwidth}
        \centering
        \includegraphics[width=\linewidth]{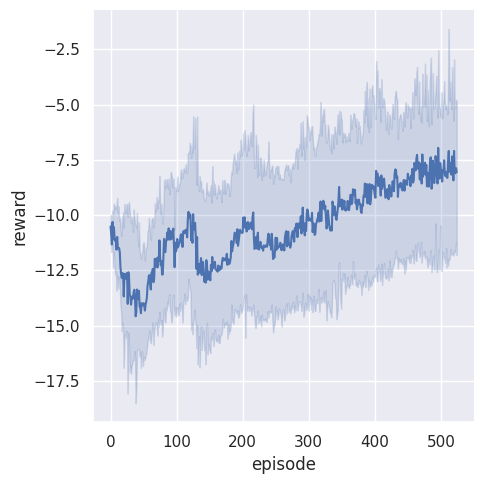}
        \caption{Beta policy learning curve.}
        \label{fig:quadcopter_beta_learning_curve}
     \end{subfigure}
     \hfill
     \begin{subfigure}[t]{0.49\textwidth}
        \centering
        \includegraphics[width=\linewidth]{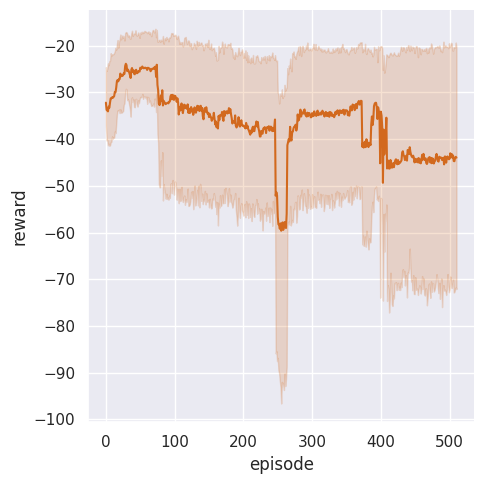}
        \caption{Projected Gaussian learning curve.}
        \label{fig:benchmark_learning_curve}
     \end{subfigure}
     \vspace{-1mm}
        \caption{Learning curves corresponding to experiments pictures in Figures \ref{fig:quadcopter_experiments_beta} and \ref{fig:phase_3_bench}. Curves show means and 95\% confidence intervals over 6 independent replications. Our CBF-constrained Beta policies clearly learn to improve reward and eventually find the goal, while the projection-based approach fails.}
        \label{fig:learning_curves}
\end{figure*}

Learning curves for the experiments illustrated in Figures \ref{fig:quadcopter_experiments_beta} and \ref{fig:phase_3_bench} are presented in Figures \ref{fig:quadcopter_beta_learning_curve} and \ref{fig:benchmark_learning_curve}.

\section{Computing Resources}
We ran our experiments on both a personal laptop and an HPC cluster. The laptop was configured with a  6-core  i7-8750H, 2.20GHz CPU, an NVIDIA GeForce RTX 2070 GPU, and  32GB RAM . The HPC server node was configured with a 32-core  Intel Xeon CPU , an 80GB  Nvidia Tesla  GPU, and  512 GB RAM .